\documentclass[conference,table,xcdraw]{IEEEtran}
\IEEEoverridecommandlockouts
\usepackage[numbers,sort]{natbib}

\usepackage{amsmath,amssymb,amsfonts}
\usepackage{algorithmic}
\usepackage{graphicx}
\usepackage{textcomp}
\usepackage{xcolor}

\usepackage{xspace}    
\usepackage{xcolor}
\usepackage{paralist}
\usepackage{url}

\usepackage{booktabs}

\newcommand{\eg}{e.\,g.,\xspace}
\newcommand{\ie}{i.\,e.,\xspace}
\newcommand{\cf}{cf.\xspace}

\newcommand{\Ie}{I.\,e.,\xspace}
\newcommand{\andor}{\textsc{ANDOR}\xspace}
\newcommand{\eat}[1]{}

\usepackage{makecell}
\usepackage{rotating}
\usepackage{colortbl}
\usepackage{multirow}
\newcommand{\myrotcell}[1]{\rotcell{\makebox[0pt][l]{#1}}}

\begin{document}

\title{Saliency Maps are Ambiguous: Analysis of Logical Relations on First and Second Order Attributions   
}

\author{\IEEEauthorblockN{Leonid Schwenke}
\IEEEauthorblockA{\textit{Semantic Information Systems Group (SIS)} \\
\textit{Osnabr\"uck University},\\
Osnabr\"uck, Germany \\
leonid.schwenke@uni-osnabrueck.de}
\and
\IEEEauthorblockN{Martin Atzmueller}
\IEEEauthorblockA{\textit{SIS Group}, \textit{Osnabr\"uck University} \&\\
\textit{German Research Center for Artificial Intelligence}\\
\textit{(DFKI)}, 
Osnabr\"uck, Germany \\
martin.atzmueller@uni-osnabrueck.de}
}

\maketitle

\begin{abstract}
Recent work uncovered potential flaws in \eg attribution or heatmap based saliency methods.
A typical flaw is a confirmations bias, where the scores are compared to human expectation. Since measuring the quality of saliency methods is hard due to missing ground truth model reasoning, finding general limitations is also hard. This is further complicated, because masking-based evaluation on complex data can easily introduce a bias, as most methods cannot fully ignore inputs. In this work, we extend our previous analysis on the logical dataset framework ANDOR, where we showed that all analysed saliency methods fail to grasp all needed classification information for all possible scenarios. Specifically, this paper extends our previous work using analysis on more datasets, in order to better understand in which scenarios the saliency methods fail. Further, we apply the Global Coherence Representation as an additional evaluation method in order to enable actual input omission.

\end{abstract}

\section{Introduction}

In eXplainable Artificial Intelligence (XAI), a lot of heatmap-based interpretation techniques emerged to produce so-called saliency maps. These aim to highlight the most class relevant features -- sometimes also referred to as attribution-based methods~\cite{adebayo2018sanity, ju2021logic}. However, a lot of work identified flaws for those methods, \eg using adversarial attacks.
But, \citep{ju2021logic} showed that many such approaches are often not helpful and can be misleading. Additionally, they emphasize the need for ground truth explanation data, to avoid a confirmation bias.
Considering that for the same model, different saliency-based methods can output completely different scores~\cite{turbe2023evaluation}, one central question arises: What do those scores actually mean?

In \citep{Schwenke2024SaliencyMA} we showed that all analysed saliency-based methods failed to capture all relevant information and even tend to encode class relevant information in the ordering of the scores -- in contrast to the expectation that high values are class relevant. We used a logical dataset called ANDOR, to approximate possible ground truth reasoning data, while using a non-informative input baseline as reference. In this work, an adapted and significantly extended revision of~\cite{Schwenke2024SaliencyMA}, we further substantiate our claims: We (1) extend the experimentation with more basic logical data sets (focusing on one or two logical gates), (2) analyse additional metrics, and (3) rank the results based on logical scenarios to give a contextualized in-depth view on the performance of saliency methods.  

Additionally, we argue that if saliency scores approximate importance, there must a method which can use a (sub)set of inputs and their saliency scores to perform classification, \eg similar to how weights are used for interpretation in linear combinations. Otherwise, the scores are ambiguous and their trustworthiness questionable. While this claim is general, in this work we propose and apply the Global Coherence Representation (GCR) framework~\cite{SA:21:global, schwenke2021abstracting, schwenke2023extracting} for interpretation and evaluation. Due to the limitations of first order assignments (one attribution score per input)~\cite{molnar2020interpretable, harris2021joint, kumar2021shapley, tsai2023faith, konig2024disentangling} the need for higher order interaction scores was recently emphasized: On the aggregation of input interactions into one score per input, important information is lost. For \eg SHAP-based approaches, multiple new methods were introduced to tackle those problems, approximating higher order interaction values~\cite{fumagalli2024shap, harris2021joint, tsai2023faith, zhang2021interpreting, tsang2020does, janizek2021explaining}. We thus extend our experimentation towards second order attribution scores, \ie showing how different inputs interact with each other, \cf GCR.

Our contributions are summarized as follows:
\begin{itemize}
    \item We extend our experimentation in~\cite{Schwenke2024SaliencyMA} with two new types of logical datasets, two new saliency methods and use further new metrics for providing a more comprehensive analytical view on the analysed methods. 
    \item We investigate three different ways to interpret numeric saliency scores (per method) as relevancy representation.
    \item We broaden our logical analysis towards specific logical scenarios and rank each saliency method according to performance to differentiate which saliency method performs better in which scenario. 
    \item We further extend our experimentations with second order attribution scores, showing practical advantages of second order attribution to differentiate logical relation.
    \item We argue for the need of a weight-based interpretation of attribution scores using simple methods that can ignore inputs, \ie proposing and evaluating the GCR~\cite{SA:21:global, schwenke2021abstracting, schwenke2023extracting} 
\end{itemize}

The remainder of this work is structured as follows: Section~\ref{sec:related} discusses related work. After that, Section~\ref{sec:background} discusses background, while Section~\ref{sec:experimentation} outlines our experimental setup. Next, Sections~\ref{sec:results}-\ref{sec:discussion} present, discuss and reflect on our results. Finally, Section~\ref{sec:conclusions} concludes with a summary and outlines interesting directions for future research.

\section{Related Work}\label{sec:related}

While many concepts for explainability exists (\eg some are summarized by \citep{rojat2104explainable, carvalho2019machine}), finding a numeric evaluation metrics is not trivial as a model's reasoning cannot be extracted that easily, \ie the model ground truth reasoning is often missing~\cite{ju2021logic}. A few numeric metrics are summarized by \eg \citep{li2021experimental, carvalho2019machine}.
A major challenge in XAI is currently that while multiple saliency methods exist, it is hard to tell which one is the best in which use-case. Here, multiple approaches ~\cite{ju2021logic, adebayo2018sanity, kindermans2019reliability, shah2021input, kokhlikyan2021investigating} already found flaws in saliency methods, based on different criteria. Some work even shows the insensitivity of many methods to the input, approximating thus general edge detectors~\cite{adebayo2018sanity,sixt2020explanations}. Further, \citep{ju2021logic} showed that many approaches for qualitative evaluation run into logical reasoning traps, when ground truth data is not available. One example is the confirmation bias, where a reasoning seems plausible, just because it could be expected. However, multiple valid ways can exist to solve a specific task, even it may seem unintuitive for humans, \ie not being plausible~\cite{jacovi2020towards}.

Another challenge in XAI is that neural networks cannot fully ignore inputs~\cite{haug2021baselines, sturmfels2020visualizing}. For this reason, re-train model approaches like \eg RemOve And Retrain (ROAR) \citep{hooker2019benchmark} were suggested. However, masking inputs in the data can induce undesired information as well~\cite{rong2022consistent,parkgeometric, Schwenke2024SaliencyMA}. For example, \citep{rong2022consistent} demonstrated this on image data, where shapes and redundant information are very common. In \citep{Schwenke2024SaliencyMA} we further extended this by showing that all analysed saliency methods did at least to some extent encode class relevant information, in a more complex manner than indication by image contours. Hence, we argue for the use of methods that can fully ignore inputs for interpretation purposes and thus include the GCR~\cite{SA:21:global, schwenke2021abstracting, schwenke2023extracting}, as an additional verification tool that can interpret any order of saliency scores as class relevancy weights.

Second order interactions/saliency scores in general are nothing new, \eg \citep{molnar2020interpretable} gives a good overview on this topic. However, second order interactions for DL are quite understudied~\cite{cui2020learning}. Recently, first order limitations of especially SHAP have been further analysed~\cite{slack2020fooling, fumagalli2024shap, kumar2021shapley, fryer2021shapley, harris2021joint, konig2024disentangling}, showing how correlations and class relevant interactions between variables are lost when they are aggregated into first order scores. Thus, many new higher order interaction methods started to emerge~\cite{fumagalli2024shap, harris2021joint, tsai2023faith, zhang2021interpreting, tsang2020does, janizek2021explaining}. For this reason, we include SHAP-IQ~\cite{fumagalli2024shap}, a unified approximation algorithm for any-order interactions, for our second order validation. Further, self-attention naturally also supports to some extent interactive properties~\cite{janizek2021explaining, SA:21:local, SA:21:global, schwenke2023extracting}, which is why we also include it into our second order experiments.

Using logical operations or reasoning for DL interpretation is nothing new~\cite{zhang2021survey, marques2023logic, fan2017revisit}, sometimes also applied as approximation on symbolic networks~\cite{garcez2022neural, srinivasan2019logical}. \citep{zhang2021survey} and \citep{pedreschi2019meaningful} even highlight the need of logical reasoning for better explanations, thus showing the advantage to approximate logical reasoning with attribution techniques.

\cite{yalcin2021evaluating} tried to verify the necessity of certain inputs in a logical formula dataset using attribution scores of non-DL methods. However, they only captured the dataset ground truth and failed to capture the model reasoning, due to suboptimal model performance, not using all possible input combinations and neglecting statistical structures. In contrast,~\cite{tritscher2020evaluation} used a logical dataset with a 100\% accurate DNN to verify saliency scores. However, they only approximated the maximal information coverage and thus mistreated redundant information, compared to~\cite{yalcin2021evaluating}. In \citep{Schwenke2024SaliencyMA} we considered those flaws using a DNN, by including a relative non-information attribution score baseline, by differentiating between maximal and minimal information coverage and by providing an exhaustive analysis including more metrics. While also focusing on the logical information characteristics of the logical framework (\andor), we introduced an approach to approximate all possible model reasonings in a controlled experiment environment.

In this paper, we further extend our analysis on the \andor dataset framework~\citep{Schwenke2024SaliencyMA}, by using more datasets and focusing on specific relation settings to rank  methods in different scenarios. We further include second order saliency methods and do a weight-based score analysis using the GCR as interpretation and verification method for global consistencies.

\section{Background and Assumptions}\label{sec:background}
In this section, we include necessary background information and assumptions relying on the formalization in~\cite{Schwenke2024SaliencyMA}. We extend them appropriately, discuss challenges in saliency map evaluation, and introduce the GCR framework.

\subsection{Information Flow}
\label{sec:informationflow}

In the following, we introduce formal notation for our task.
We consider a function $f\colon D \rightarrow C$ (representing a classification task $T$), given a finite set of classes $C$. Let $D$ denote a dataset that captures all possible instantiations with respect to a given feature set. Thus, $D$ contains all possible data samples $d\in D$ of length $l$, with $d=(d_1, \dots, d_l)$ being a tuple of input (features). An input $d_j \in d$ at position $j$ is thus taken from the finite (universal) input domain $M$, \ie $d_j\in M$ and $D = M^l$. 
For deriving/explaining a class $c \in C$ for a specific $d \in D$ (with $f(d) = c$), however, often only a subset of inputs $s \subseteq d$ is required.
A class $c \in C$ is a set of hidden information $c$ represented by sets $c_i \subseteq H$, considering a universal set $H$ of all hidden class discriminative information of any form, \eg also dataset distribution or implicit information; \ie each $c_i$ represents a distinct valid way to derive class $c$. Because each dataset can have unique class discriminative decisions to derive a class, we combined all possible decision information under the term hidden information; \eg a specific decision tree or a logical formula indicates a possible decision processes $c_i$. However, this does not mean that no other valid decision criteria exists.
In our context, this means that a function $f^D$ for $T$ exists that can extract a set of hidden information for each input $d_j \in d, d \in D$. We define this as $f^D\colon D \times \{1, \dots, l\} \rightarrow Q$, with $Q \subseteq H$. Modelling all $q \in f^D(d, j)$ however is often quite hard, because this relates to knowing all possible ways to derive a class. The function $f^D$ is subject to two important constraints: (1) $\forall d \in D\colon (f(d) = c^1) \rightarrow \exists c_i \in c^1\colon c_i \subseteq \bigcup\limits_{d_j \in d} f^D(d,j)$, \ie the hidden information indicated by the union of all $d_j \in d$ must contain the needed information for at least one $c_i \in c^1$. (2) Each $d \in D$ contains only the information set of exactly one class:  $\forall c_j \in c^2 \in C\colon c_j \subseteq \bigcup\limits_{d_j \in d} f^D(d,j) \rightarrow c^1 = c^2$. With $f^D$ we can derive a set of all $s \subseteq d$ that contain the information needed for $f(d) = c$, defined as $R^d = \{s \mid \exists c_i \in f(d)\colon c_i \subseteq \bigcup\limits_{d_j \in s} f^D(d,j) \wedge s \subseteq d\}$, assuming $s$ keeps the position information of $d$. $R_{min}^d = \{r_1 \mid r_1 \in R^d \wedge \neg \exists r_2 \in R^d\colon |r_2| < |r_1|\}$ is the minimal information coverage for input $d$, \ie all minimal sets of relevant features to derive class $f(d) = c$. $R_{max}^d = \{r \mid r\in R^d \wedge \forall d_j \in r \colon \exists c_i \in f(d) \colon f(d, j) \cap\ c_i \neq \varnothing \}$ is the maximal information coverage of $d$, \ie including all relevant inputs that contain class relevant information. All other inputs are irrelevant for class $f(d) = c$. Those definitions describe our goal, \ie to find a complete estimation on $R^d_{min}$ and $R_{max}^d$ (all possible way to derive a class) for the simple dataset \andor, as described below. Thus, it is possible to check if the suggested saliency score ranking/ordering matches one possible reasoning $r\in R^d_{min}$. For this, we later introduce a non-informative input baseline, to differentiate between relevant and non-relevant inputs in different scenarios.

\subsection{Background and Challenges on Saliency Maps Evaluation}
\label{sec:assumptions}

As interpretability and explainability are quite vaguely defined, \cf~\citep{zhang2021survey}, assumptions for saliency maps are either not very clearly defined as well or only describe partially what they actually mean~\cite{zhang2021survey}.
With many unclear assumptions, the question arises, how to actually verify the quality of a saliency map? A typical idea is that each input gets a local score assigned, which represents its relevancy/contribution towards the output class. Therefore, a popular way of verification is by ablating inputs (often using masks) based on the score ranking/distribution~\cite{hooker2019benchmark, shah2021input, rong2022evaluating,sturmfels2020visualizing}. This is done by removing either a percentile of the highest (MoRF -- Most Relevant First) or lowest scored inputs (LeRF -- Least Relevant First) \citep{tomsett2020sanity}. For LeRF the accuracy is expected to be maintained as long as possible, while for MoRF a fast accuracy drop is expected. After the stepwise removal of data, typically one of several Area Under the Curve (AUC) metrics is calculated~\cite{gomez2022metrics, tomsett2020sanity, turbe2023evaluation}, for measuring the effectiveness of the removed information. Based on this approach and the implied assumptions, we list a few challenges for saliency evaluation and interpretation:

\paragraph{Redundancy}
An important but often overlooked differentiation is if the saliency method does capture all classification information or just a minimal set of information (\cf difference between $R^d_{min}$ and $R^d_{max}$). For a minimal set and the MoRF strategy, this would result in a less steep reduction, as redundant but class relevant information is included in the not removed data. For this reason, we focus on a LeRF strategy, to better capture redundant information, as the model performance should be maintained as long as possible if the saliency method has a good/valid score distribution. 

\paragraph{Non-Informative Baseline}
Another unclear question is if a score of 0 is really non-informative, or which score stands for uninformative information. For this reason, we included an input baseline that is output irrelevant (\ie does not include any class relevant information) and hence should not be scored as relevant. With this, a more dynamic score baseline for non-informative inputs is enabled in our experiments, to avoid this problem. As already presented in~\cite{Schwenke2024SaliencyMA}, many methods assigned the irrelevant inputs a score higher than zero or even a strongly varying score, thus verifying this apprehension. 

\paragraph{Linearity}
It is unclear if the relevancy scoring is linear, \eg input $d_1$ is scored with $8$, $d_2$ with $4$ and $d_3$ with $2$, the question is if $d_1$ is double as relevant as $d_2$ and quadruple relevant as $d_3$ for the output class or if this distances are not linear. This is hard to test for, which is why we expect a more or less linear distribution for the scores, as it is a desirable property and often indirectly assumed. In contrast to typical approaches where the same fixed percentile of inputs is removed for all samples, \cf~\cite{hooker2019benchmark, shah2021input, rong2022evaluating}, our evaluations are primarily done on the ranking of saliency scores by comparing them towards the non-informative input baseline. Thus, this property is only relevant for our global evaluation.

\paragraph{Global Comparability}
As saliency methods are primarily locally focused, the aggregation towards global explanation is not necessary given~\cite{saleem2022explaining}, \eg if a score of 5 from one sample is equivalent to a score of 5 from another sample. This property is very important for finding dataset-wide relations between inputs, \ie the dynamic global interactions between features. As we include a baseline for relativization, this property is only relevant for our global analysis. In~\cite{Schwenke2024SaliencyMA} the high standard deviation indicates that this property is probably only somewhat given, \cf stability~\cite{li2021experimental}. We therefore assume at least somewhat comparable values between samples, \ie an important score should on average still be highly scored, even with slight deviations.

\paragraph{Ablating Inputs}

A general challenge in XAI is how to ignore inputs without introducing unwanted information~\cite{rong2022consistent, haug2021baselines, sturmfels2020visualizing, parkgeometric, Schwenke2024SaliencyMA}. As neural networks cannot fully ignore inputs, masked are used. However, the influence of masked input is often unclear~\cite{sturmfels2020visualizing}. \citep{hooker2019benchmark} showed that masking inputs on a trained model to verify if the model still outputs the same conclusion often runs into a distribution shift. Thus, they proposed a \textit{RemOve And Retrain} (ROAR) approach, \ie training a new model with the masked data to verify if classification relevant information is still included. This led to another challenge, as masked values could encode classification information. For example, \citep{rong2022consistent} demonstrated this on image data, where shapes and redundant information are very common. In \citep{Schwenke2024SaliencyMA} we further extended this by showing that all analysed saliency methods did at least to some extent encode class relevant information, in a more complex manner than indication by contours. Some methods tried to minimize those effects by \eg interpolation~\cite{parkgeometric} or augmented training processes~\cite{turbe2023evaluation}. However, those approaches are rather domain specific or not appliable for our controlled experiments. Anyhow, it is still unclear if fully ignoring inputs for neural networks is possible~\cite{sturmfels2020visualizing}. 
To be able to ignore inputs, we further extend the analysis from~\cite{schwenke2023extracting} by using the GCR~\cite{SA:21:global, schwenke2021abstracting, schwenke2023extracting}, a symbolic global weighted occurrence based classification approach, to approximate simple model reasoning in an input symbol $\times$ saliency score per position manner. The GCR enables a clear, interpretable weighting for each input, representing its global influence towards the different classes. Herby, a first and second order interaction variation exists, which showed promising results for the model Fidelity~\cite{carvalho2019machine} on time series data and thus can help to build a great foundation for logical pattern detection and analysis.

\subsection{Assumptions on Saliency Maps}

Based on those evaluation procedures and the logical information flow (see Section~\ref{sec:informationflow}), we consider the following basic assumptions/expectations:

\begin{enumerate}[(A)]
    \item The ranking order between scores is relevant, \ie an input with a higher saliency score than another input is more relevant in the model's local classification process. Therefore, a lower scored input has less, no, or even class contradicting information (w.r.t the model's decisions).
    \item If two inputs contain redundant information, at least one input needs to be relevant (\ie as described above).
    \item The saliency score ranking should not depend on independent inputs that do not contribute to the output.
    \item Dataset distributions and implicit information (\eg shape of masked inputs) can contribute towards information which is relevant for the class.
\end{enumerate}
Based on those assumptions above, we formulate additional, more specific assumptions for our task: 
\begin{enumerate}[(A)]
    \item[(E)] Class irrelevant inputs $d_j \in d,\, d \in D$ act as a (non-informative) input baseline, with $\forall r \in R^d_{max} \colon d_j \not\in r$. Hence, such a baseline input $d_j$ should not have a higher saliency score than any class relevant input $d_k \in r \in R^d_{min}$ (see Assumptions A, B and C).
    \item[(F)] If the saliency metric approximates classification information (Assumptions A, B and C), for a logical dataset, then the logical accuracy should be preserved as long as possible due to the input independence and clear $R_{min}$.
    \item[(H)] Correlations between saliency scores are undesired if the inputs are logically independent or class opposing (assumption D).
\end{enumerate}

Because we additionally focus in this work on (global) interactions, we further assume:
\begin{enumerate}[(A)]
    \item[(I)] The value of a saliency score for a specific class, should have a value-wise relative comparability towards other samples, \ie if two inputs from two different samples have the same score, they should be similarly ``important''.
\end{enumerate}
If Assumption I would not be included, then the saliency scores would be inconsistent between samples and thus, the extraction of global interactions or relations is questionable. This inconsistency is in general undesirable, as saliency scores can easily be misinterpreted and thus always reference points would be needed for interpretation.

\subsection{GCR}
\begin{figure}[ht!]
	\centering
	\includegraphics[width=0.95\columnwidth]{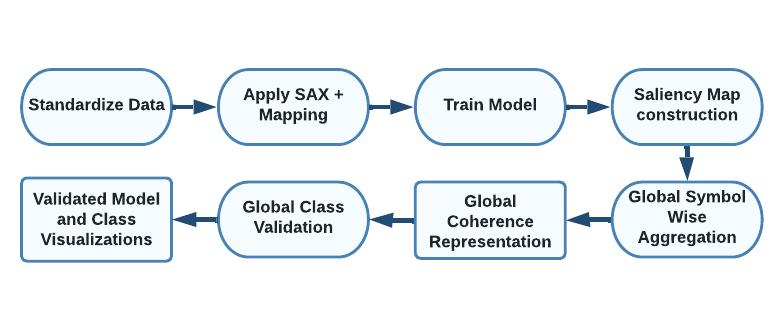}
	\caption{The process pipeline of our Global Coherence Representation, \cf~\cite{schwenke2023extracting}.}
	\label{fig:globalpipeline}
\end{figure}

\begin{figure}[h!]
	\centering
	\includegraphics[width=0.95\columnwidth]{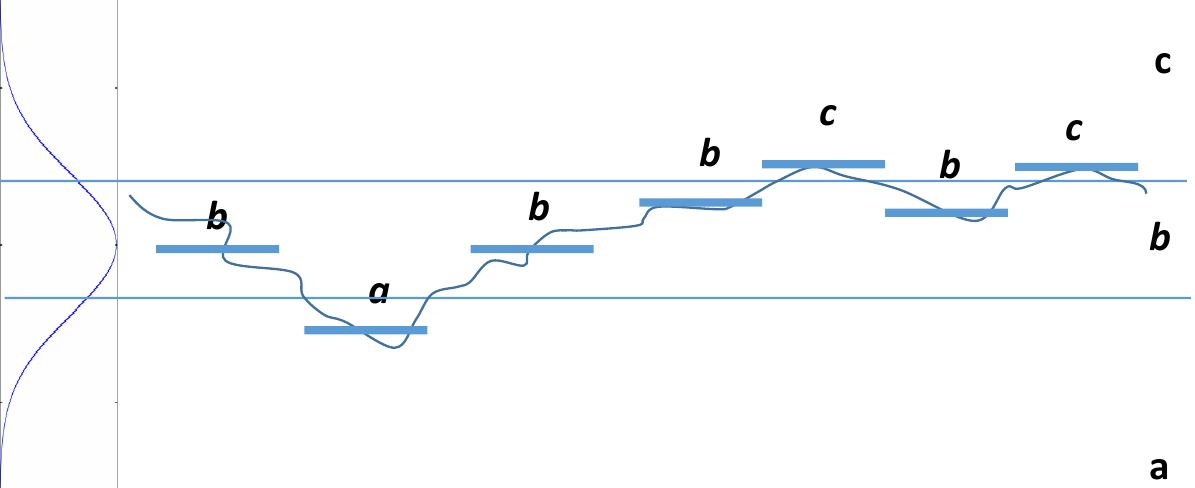}
	\caption{Example visualisation for a SAX discretization, \cf~~\cite{atzmueller2017explanation}: Each data point from the original time series is mapped to a discrete symbol (a, b, c) based on the quantiles from the standard normal distribution.}
	\label{fig:sax}
\end{figure}

\begin{figure}[ht!]
	\centering
	\includegraphics[width=1\columnwidth]{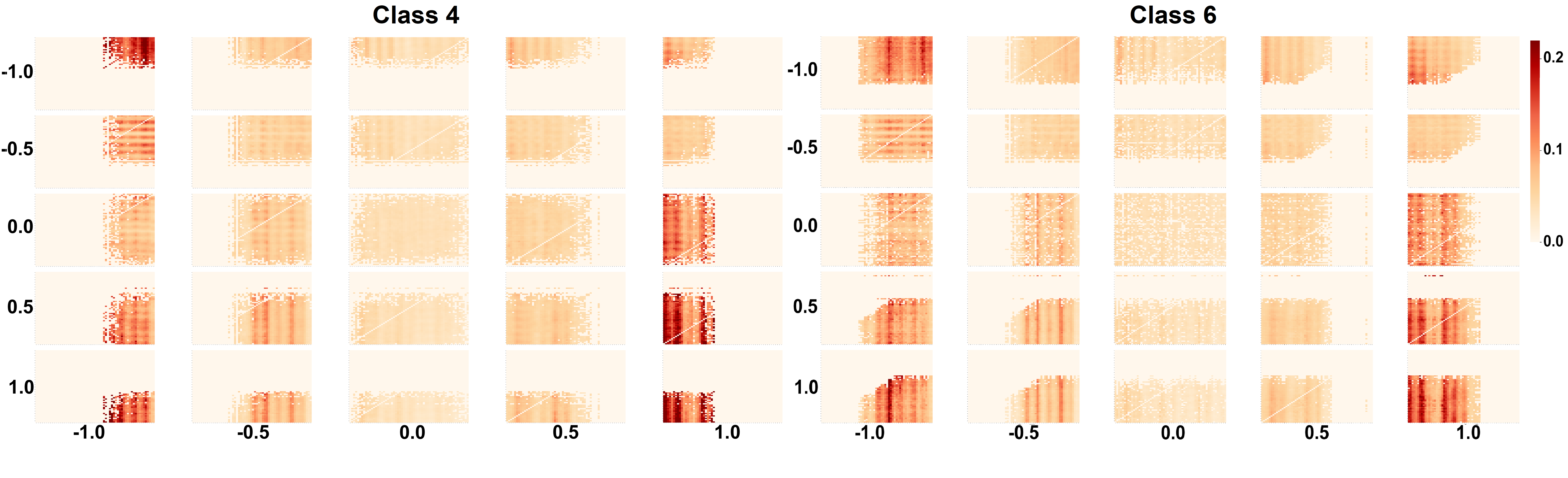}
	\caption{Two FCAMs from the Synthetic dataset, w.r.t. class 4 (left) representing a slowly falling trend and class 6 (right) representing a sudden falling trend, \cf~\cite{SA:21:global, schwenke2023extracting}.}
	\label{fig:fullClassBoth}
\end{figure}

\begin{figure}[ht!]
	\centering
	\includegraphics[width=1\columnwidth]{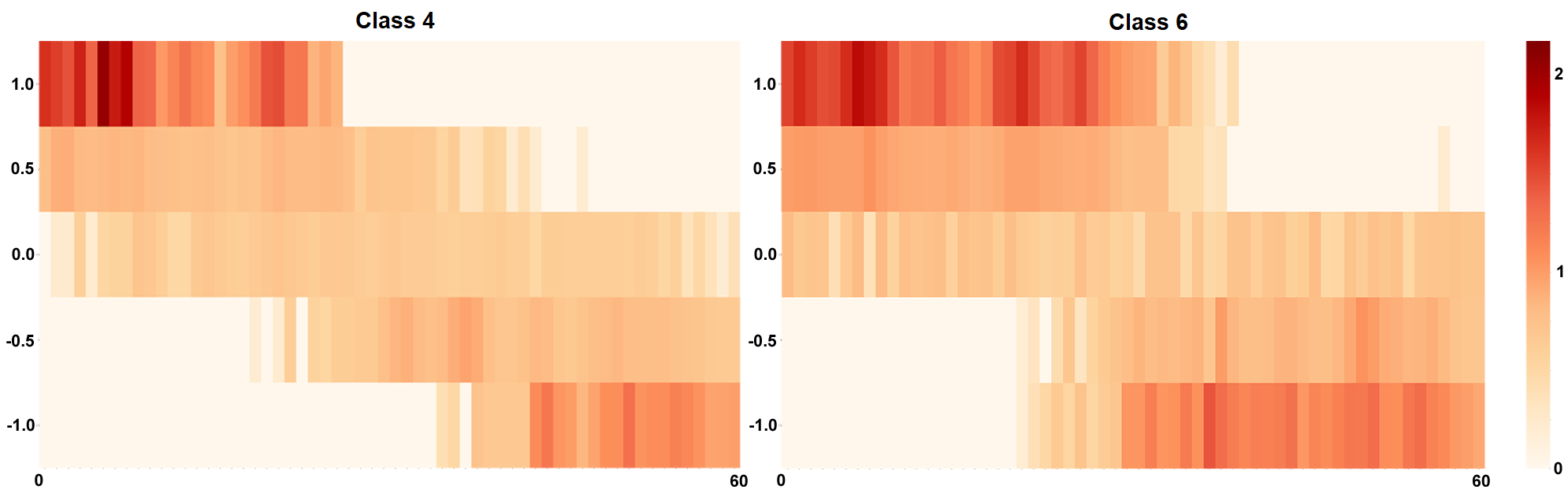}
	\caption{Two GTMs from the Synthetic dataset, w.r.t. class 4 (left) representing a slowly falling trend and class 6 (right) representing a sudden falling trend, \cf~\cite{SA:21:global, schwenke2023extracting}.}
	\label{fig:minClassBoth}
\end{figure}

The Global Coherence Representation (GCR)~\cite{SA:21:global, schwenke2023extracting} is a representation/model which uses saliency scores for symbolic interpretable approximation on a task or model level, \ie approximating a target reasoning. In the GCR a global score per symbolic input value per input position indicates the class affiliation of this specific input; enabling membership classification. Figure~\ref{fig:globalpipeline} shows the pipeline of the approach. The first step standardizes the data, the second step symbolizes the data using Symbolic Aggregate Approximation (SAX)~\cite{Lin:2003,LKWL:07,rojat2021explainable} into $v$ many symbols. SAX is an aggregation and abstraction technique in the area of time series analysis, \eg also enhancing interpretability and computational sensemaking. Figure~\ref{fig:sax} gives a small example of how each symbol is extracted. Hence, our model is trained with a finite input dimension vocabulary, more similar to how NLP inputs work. In the second step, the symbolic inputs get mapped into $v$ equally distributed numbers from [0,1], based on the numeric order of each symbolic bin. For example, four symbols would be represented by \{-1,-0.3333, 0.3333,1\}. The third step is training a model on this symbolic data, and the fourth step is extracting and if necessary aggregating the attribution scores into the desired format. In the fifth step, all local attribution scores are aggregated over the occurrence of a specific symbol or a pair of symbols at a specific position towards a specific class. In consequence, the GCR is represented by a set of matrices (from each possible input -- to each possible input) per class, that highlights how strongly scored (on average when occurring) a specific input at a specific position is (for each class). Figure \ref{fig:fullClassBoth} and Figure \ref{fig:minClassBoth} show a GCR example each, for the first order and second order aggregation. The last steps are the validation processes, as the GCR can be used for classification. 

We focus on the two main GCR models, the FCAM and the GTM. The GTM is a first order method to enable a simple split of (average) scores between each possible symbolic value at each position, \ie showing a simple global range of possible values per class. An example for the GTM can be seen in Figure \ref{fig:minClassBoth}, where two classes from a trend-based dataset are compared. The left representation showing a slow trend drop, while the right one a more sudden one. The FCAM is a second order approach which uses the symbol-to-symbol scores at each position for each symbol to give a more detailed overview. Figure \ref{fig:fullClassBoth} shows the same example as for the GTM. Hereby, each matrix represents how each symbolic value interacts with each other, where the most bottom left corner of each matrix represents how strong the first inputs highlights itself. In summary, the GCR shows how strong the occurrence of a symbol/symbol-pair effects the output and is thus easy to interpret. For more details regarding the representations or the different processes, we refer to the presentation in~\citep{schwenke2023extracting}.

\paragraph{GCR Classification and Membership Function}

The basic principle behind the applied GCR classification method is based on the idea, that each score for each symbol/symbol-pair at each position is summed up into a membership score. This means, that each position is seen as a weighted contribution towards a specific class. For normalization, each score is divided by the maximal reachable score per class in order to enable a percentile membership comparison.

The GTM classification approximates the simple first order membership function:
\begin{equation}S^c(X) = \frac{\sum\limits_{i=1}^n  P(x_i \vert c, i) }{ \sum\limits_{i=1}^n  \max\limits_{v \in V}(P(v \vert c, i))} \end{equation}\\
with $n$ denoting the length of the time series, a given class $c$, vocabulary $V$, $d_i \in d$ being the $i$-th element in sequence $d$ and function $P$ providing the probability of the given conditions.
The specific GTM membership-function uses attribution scores for weighting:\\
\begin{equation}S^c_{GTM}(X) = \frac{\sum\limits_{i=1}^n  P^{x_i-c}_i}{\sum\limits_{i=1}^n \max\limits_{v \in V}(P^{v-c}_i)}\end{equation}\\
with $P^{v-c}_i \in $ GTM being our GTM approximation of $P(v \vert c, i)$.\\
The extended general second order membership-function is:
\begin{equation}
S^c_{2D}(X) = \frac{ \sum\limits_i^n\sum\limits_j^n  P(x_i,y_j \vert c, i,j) }{\sum\limits_i^n\sum\limits_j^n  \max\limits_{u \in V} \max\limits_{v \in V}(P(v,u \vert c, i,j))}\end{equation}\\
And therefore the FCAM membership-function approximates this via:
\begin{equation}S^c_{FCAM}(X) =  \frac{\sum\limits_i^n\sum\limits_j^n  P^{x_iy_j-c}_{ij} }{\sum\limits_i^n\sum\limits_j^n  \max\limits_{u \in V} \max\limits_{v \in V}(P^{uv-c}_{ij})}\end{equation}\\

In order to illustrate how the classification works, we can consider the different GCRs as reward distributions: The further the values are from the peak at each position, the less the input is part of the class. Assuming, a specific feature $d_i$ with value 1 is on average important (high scored), while the same feature $d_i$ with value -1 is scored less high. Given our assumptions: the contribution of $d_i$ with value 1 to a class $c$ is higher than $d_j$ with value -1 (\cf Assumption A) and thus the occurrence of the feature $d_i$ with value 1 makes the class $c$ more likely than the feature $d_i$ with value -1. We argue that given our assumptions, such weighted interpretation should be possible and is desirable for evaluation and interpretation of attribution based methods. As the order of saliency scores is relevant to reduce aggregation-based problems~\cite{slack2020fooling, fumagalli2024shap, kumar2021shapley, fryer2021shapley, harris2021joint, konig2024disentangling}, FCAM provides unique insight into second order saliency scores.

\subsection{Upscaling 1D to 2D}
\label{sec:upscaling}
Not all saliency methods support second order scores. Nevertheless, we want to upscale all first order methods to show the possibility of extracting global relation information from aggregated scores. For this, we consider the values of the first order dimension $l$ (the length of the sample), and construct a $l \times l$ matrix by duplicating the values $l$ times: With this, all values are put into relations with all other values at any other position. Given a global interpretation, pairs of inputs which act together will get higher global scores, as both inputs have high scores when co-occurring.

\subsection{Datasets}

\begin{figure}[ht!]
	\centering
	\includegraphics[width=0.74\columnwidth]{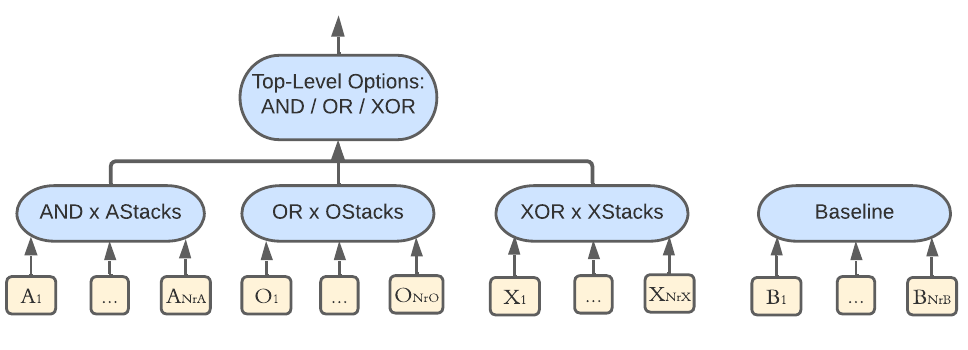}
	\caption{Framework for the \andor dataset.} 
	\label{fig:andor}
\end{figure}

To narrow down our expectations on valid saliency maps, in~\cite{Schwenke2024SaliencyMA} we constructed the dataset-framework \andor. An \andor dataset is based on propositional logical operators, modelling a two layered relation between operators. The first layer of the logical formula is described by four different blocks (representing \textit{AND}, \textit{OR}, \textit{XOR} and \textit{Baseline}), where each block contains individually many logical gates (\textit{AStacks, OStacks, XStacks}), for which each gate has a fixed predefined length per block (\textit{NrA, NrO, NrX, NrB}). \textit{XOR} is hereby defined as \emph{true} when having exactly one \emph{true} (positive) input. The baseline block acts as reference for non-informative inputs, \ie relevant inputs should be higher scored than the scores that the baseline receives (Assumption E). Each gate input takes on a value from the domain $M$, limiting all possible input values. Here, a binary case with $M =\{0,1\}$ is the simplest one. To be able to evaluate each binary logical operation, a set of positive representations $T \subseteq M$ is given, where all values $m \in M$ with $m \not \in T$ are handled as a negative gate inputs. The final output is decided by the respective \textit{AND}, \textit{OR} or \textit{XOR} gate (top-level). While non-binary outputs would be possible, for simplicity and clarity we only focus on the binary outputs (True/False or 1/0), which can naturally occur in more complex sub-settings. Figure \ref{fig:andor} illustrates this concept. To analyse different situations, we deploy 21 different \andor datasets for our experiments (12 more than in~\cite{Schwenke2024SaliencyMA}), based on the number of three different top-levels multiplied by our 7 parameter settings:

\begin{itemize}
\item \textit{2inBinary}: 2 inputs per gate, limited to 1 stack per gate-type, with M=\{-1, 1\} and T=\{1\}.
\item \textit{2inQuaternary}: similar to 2inBinary but with  M=\{-1, -0.333, 0.333, 1\} and T=\{-0.333, 1\}.
\item \textit{3inBinary}: as 2inBinary, but with 3 inputs per gate. 
\item \textit{2inBinaryDoubleGate}: Two gates with the same logical function, combined over the top-level, with M=\{-1, 1\} and T=\{1\}. This setting results with all gates to 3 different settings. The Dataset 2inBinaryDoubleGateAND would mean \textit{AND}-gates are used.
\item \textit{BinarySingleGate}: Only the top-level, \ie exactly one logic gate, having 4 inputs, with M=\{-1, 1\} and T=\{1\}.
\end{itemize}

\begin{figure}[ht!]
	\centering
	\includegraphics[width=0.99\columnwidth]{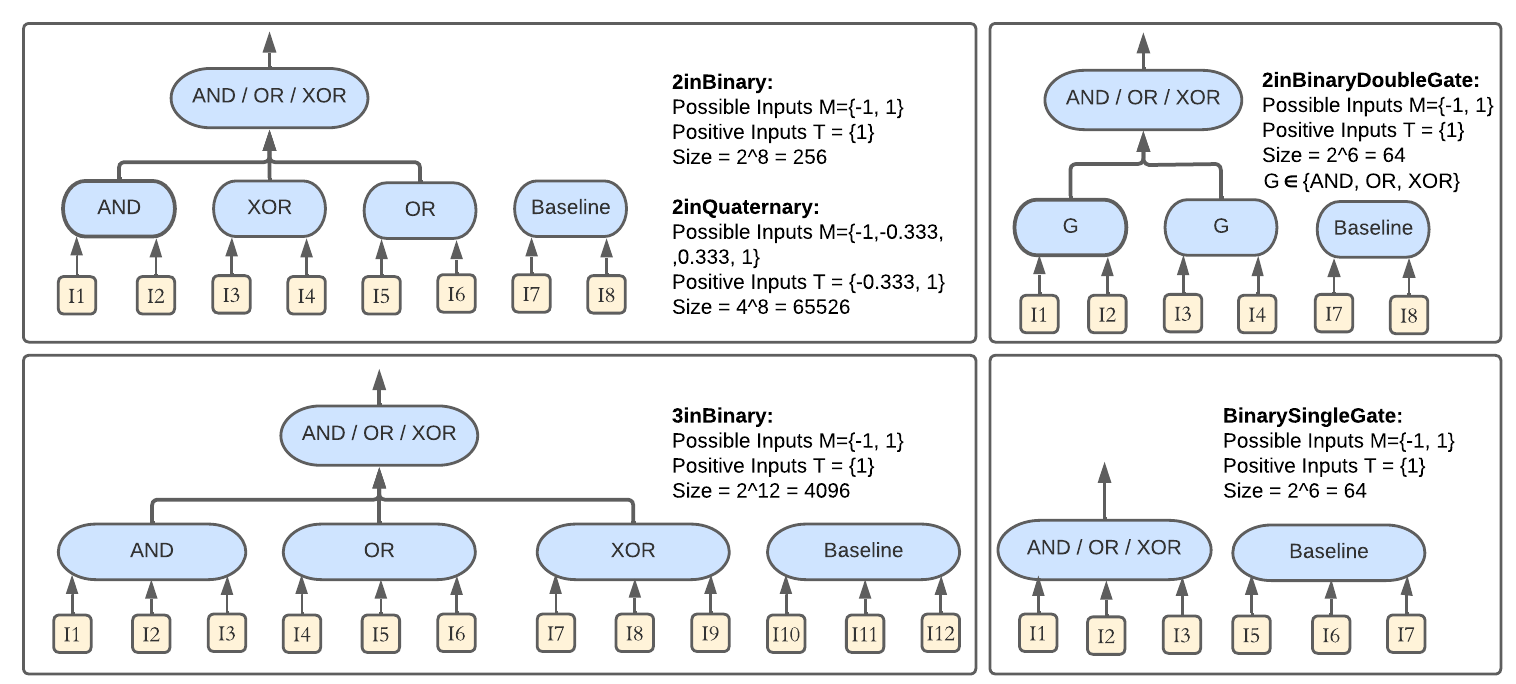}
	\caption{The five different ANDOR dataset configurations we use.} 
	\label{fig:andorDatasets}
\end{figure}

A structured example for the 5 different settings is given in Figure \ref{fig:andorDatasets}. With more \andor datasets, we aim to analyse fundamental behaviour of saliency maps in different scenarios, ranking which method performs how well under which conditions. We argue those types of analysed relations between inputs will naturally occur in more complex datasets, making \andor a good test dataset due to the fully independent inputs and clear information flow. Hence, this also enables the ability to generate all possible model reasonings, which is not feasible on typical complex datasets. 

\subsection{Expectations on \andor}

In order to make sure the model understands the task completely, we generate all possible input combinations for each \andor dataset. Hence, to reach an  accuracy of 100\%, the model needs to understand which input values per sample are relevant for each class. To minimize statistical biases (class imbalance), we balance the training set via oversampling using \emph{Imbalanced-learn}~\cite{JMLR:v18:16-365}. 
With this and due to the independence of the inputs, we minimize the effects from Assumption~$D$ and can better approximate $R^d_{min}$ per $d \in D$. By comparing all $r \in R^d_{min}$ towards the \textit{Baseline} over the whole dataset, we can find violations of Assumption~$E$. While we only look at binary outputs and in some cases binary inputs, a good explanation technique should also work for simple scenarios, also because \eg complex inputs/outputs can be sometimes clustered into binary class ranges.

The following specifies what information is expected for each logical gate:
\begin{itemize}
    \item \textbf{AND:} For $f(And_{pos}) = 1$, all values are important/complimentary $R_{min}^{And_{pos}}=\{And_{pos}\}$. For $f(And_{neg}) = 0$, only one negative input is enough (redundancy) $R_{min}^{And_{neg}}=\{d_j \mid d_j \in And_{neg} \wedge d_j \not\in T\}$.
    \item \textbf{OR:}  For $f(Or_{neg}) = 0$, all values are important/complimentary $R_{min}^{Or_{neg}}=\{Or_{neg}\}$. For $f(Or_{pos}) = 1$ , only one positive input is enough (redundancy) $R_{min}^{Or_{pos}}=\{d_j \mid d_J \in Or_{pos}\wedge d_J \in T\}$. 
    \item \textbf{XOR:} For $f(Xor_{pos}) = 1$, all values are important (exclusive information) $R_{min}^{Xor_{pos}}={Xor_{pos}}$. For a $f(Xor_{neg}) = 0$, either all inputs are important and negative or two positive inputs are important; making this case more complex $R_{min}^{Xor_{neg}}=\{Xor_{neg} \mid \forall d_j \in Xor_{neg}: d_J = 0\} \cup \{\{d_i, d_j\} \mid\ \exists d_i, d_j \in Xor_{neg} : d_i \neq d_j \wedge d_i = 1 \wedge d_j = 1\}$. 
\end{itemize}

\section{Experimental Setup}\label{sec:experimentation}

This section describes the setup of our experiments in detail.

\subsection{Models}
We use two state-of-the-art architectures with multiple parameters towards enhancing generalizability. We apply shallow networks as they perform better on sequential tasks~\cite{wen2022transformers}, also to reduce biases~\cite{haug2021baselines}.
Because multiple saliency maps like \eg GradCam~\cite{selvaraju2017grad} are mainly developed for CNNs, we use two ResNet-Blocks~\cite{he2016deep} to construct a CNN-model. Additionally, we also look into a two-layered Transformer~\cite{vaswani2017attention} to also cover Attention-based methods like \eg the Attention enhanced LRP from~\cite{chefer2021transformer}. We test out all possible saliency methods per model, while exploring the effects of certain logical structures. For details about the architectures and hyperparameters, we refer to our open source code\footnote{\label{foo:code}\url{https://github.com/lschwenke/SaliencyMapsAmbiguous}}.

\subsection{Saliency Maps}
In our experiments, we compare 14 different first order saliency methods, shown in Table~\ref{tab:methods}, \ie two more than presented in~\cite{Schwenke2024SaliencyMA}. Additionally, we analyse 4 second order methods, primarily using the FCAM. For all other methods, we upscale the first order scores as described in Section \ref{sec:upscaling}. In the implementation from~\citep{chefer2021transformer} a CLS-Token is used to reduce the saliency score to one per input. We also look into the second order version of this implementation, by not using a CLS-Token. For all second order explanations, we apply the max-operation per row, to enable single order evaluations. 

\begin{table*}[htb!]
\centering
\caption{List of all applied Saliency methods, while listing appliable models and the implementation source.}
\label{tab:methods}
\begin{tabular}{l|c|c|c}
\toprule
\multicolumn{1}{c|}{\textbf{Method}} & \multicolumn{1}{c|}{\textbf{Models}} & \multicolumn{1}{c}{\textbf{Implementation}} & 2nd Order Support \\ \midrule
LRP-Full~\cite{bach2015pixel} & Both &~\cite{chefer2021transformer} & NO\\
LRP-Rollout~\cite{abnar2020quantifying} & Transformer &~\cite{chefer2021transformer}  & YES\\
LRP-Transformer~\cite{chefer2021transformer} & Transformer &  Adapted from~\cite{chefer2021transformer}  & YES\\
LRP-Transformer CLS~\cite{chefer2021transformer} & Transformer &~\cite{chefer2021transformer}  & NO\\
Attention~\cite{vaswani2017attention} & Transformer & Aggregation \cf~\cite{schwenke2023extracting} & YES\\
IntegratedGradients~\cite{sundararajan2017axiomatic} & Both & Captum~\cite{kokhlikyan2020captum} & NO\\
DeepLift~\cite{shrikumar2017learning} & Both & Captum~\cite{kokhlikyan2020captum} & NO\\
Deconvolution~\cite{zeiler2014visualizing} & CNN & Captum~\cite{kokhlikyan2020captum} & NO\\
GradCam~\cite{selvaraju2017grad} & Both & pytorch-grad-cam~\cite{jacobgilpytorchcam} & NO\\
GuidedGradCam~\cite{selvaraju2017grad} & Both & Captum~\cite{kokhlikyan2020captum} & NO\\
GradCam++~\cite{chattopadhay2018grad} & Both & pytorch-grad-cam~\cite{jacobgilpytorchcam} & NO\\
KernelSHAP~\cite{lundberg2017unified} & Both & Captum~\cite{kokhlikyan2020captum} & NO\\
FeaturePermutation~\cite{molnar2020interpretable} & Both & Captum~\cite{kokhlikyan2020captum}& NO\\
SHAP-IQ~\cite{fumagalli2024shap} & Both &~\cite{fumagalli2024shap}& YES\\
\bottomrule
\end{tabular}

\end{table*}

\subsection{Metrics}
To find violations of our assumptions described in Section~\ref{sec:assumptions}, we use the metrics from~\cite{Schwenke2024SaliencyMA} and extend them:

\begin{compactenum}[(1)]
    \item Needed Information below Baseline (NIB): Percentage of samples $d \in D$, where at least one input $r_j \in r, r \in R^d_{min}$ is below the highest \textit{Baseline} input per $d$. This metric checks for Assumption E, \ie if a saliency method can estimate a minimally needed relevancy of inputs correctly. If $\mathit{NIB} > 0$, then $R^d_{max}$ does also not meet the assumption as well. We difference between NIB-Full: The NIB percentage over the whole test set; and NIB-Balanced: The average between the NIBs for each class. As the number of samples per class in the test set is unbalanced, the latter assures that the failure to capture information is treated fairly for each class, while the first is more sample centred.
    \item General Information below Baseline Coverage (GIB): Percentage of inputs $d_j \in d$ over all samples $d \in D$ that contain class relevant information $\exists r: d_j \in r \wedge r \in R^d_{max}$, but are scored below the highest \textit{Baseline} input per $d$. The GIB measures in percent how much class relevant information ($R^d_{max}$) in the whole dataset from did a saliency method fail to capture. We differentiate between GIB-Full and GIB-Balanced to account for the class imbalance in the test set, \cf NIB. 
    \item Logical Accuracy: The accuracy after masking the data, by using known logical truth tables, \ie combinations of undefined inputs result in undefined. With this, we check for Assumption F.
    \item Logical Statistical Accuracy: The accuracy after masking the data, by using known logical truth tables and considering probabilities for masked inputs assignments. Therefore, taking \eg an \textit{AND}-gate, if two or more inputs are masked, then the output is more likely to be \textit{false}.
    \item Full Double Class Assignments (Full-DCA): Similar to truth tables, two equal sets of inputs (ignoring irrelevant inputs) should not lead towards different results. For all samples of the test set, where the original and the retrained model output the same class: Given a threshold $t$, the count where the relevant inputs $\{ \{d_1, \dots, d_{l-NrB}\} \subseteq d \mid d \in D\}$ map towards different classes in the retrained model, \ie the decision relevant information is in the \textit{Baseline} inputs (\cf Assumption C). Considering our Assumptions, as long as relevant inputs can remain (after applying the threshold), the DCA should be 0. To compare multiple settings, we calculate the Full-DCA in percent.
    \item Minimal-DCA: This metric is similar to Full-DCA, but we compare logical gate inputs, where it is clear when this logic gate is relevant for the output class ($|R^d_{min}| = 1$). For \eg the \textit{AND}-top-level comparison per gate: The gate inputs where only this specific gate evaluates negative, to the gate inputs, where the model output is positive (\ie all gates are needed). \Ie if one input combination for one gate leads towards different model outputs, even though different gate outputs would be necessary, then the decision information is encoded into other masked inputs.
    \item To Irrelevant Correlation: As the baseline inputs do not contribute towards the output, no correlation between the baseline scores and all other scores should exist (Assumption H). For this metric the Pearson correlation~\cite{kowalski1972effects} between each score (without baseline) is calculated towards each baseline input.
    \item GCR Fidelity: As using the GCR for classification is possible, the Fidelity~\cite{carvalho2019machine} can be calculated, to check how well the GCR can predict the model outputs. Hence, the GCR Fidelity shows how well the saliency scores can act as weights of relevancy for the different classes. 
    \item tGCR Fidelity: The Fidelity of the tGCR. The tGCR is generated by ignoring certain saliency scores (thresholded) on creation. Using the highest input baseline score per sample as threshold, this shows the influence of supposedly irrelevant scores on the GCR Fidelity; tackling globally differentiable encoding.
\end{compactenum}

\subsection{Experimentation Setup}
For each parameter combination, we perform an experiment with a 5 fold-cross-validation (for the validation set). Afterwards, a validation model (\cf ROAR~\cite{hooker2019benchmark}) is retrained for each saliency method with masked inputs based on one dynamic threshold. As discussed in Section \ref{sec:assumptions}, we only considered the LeRF removal strategy, \ie all high scored values are relevant for the task. For each sample the threshold is the highest \textit{Baseline} input score per sample, \ie Assumption E. As the Full-DCA needs a threshold smaller than \textit{Baseline} per sample, so that information actually could be included in the baseline, we introduce three further baselines (we only use those for the Full-DCA): \textit{t1.0}, \textit{t0.8}, \textit{t0.5}. In contrast to~\citep{hooker2019benchmark} we take the average saliency score per sample times a factor, to enable more dynamic masking for samples where more or fewer inputs can be relevant. The number after 't' stands for the used factors, \ie t1.0 = avg. of sample $\times$ 1. 
Further, for each trained base model, a Random Forest Model~\cite{breiman2001random} is trained for comparison. Additionally, a GCR is build using the base model's saliency scores, as well a tGCR (threshold GCR) is built for each masked dataset, ignoring on construction all saliency scores below or equal to the baseline per sample. Afterwards, all our metrics are calculated. In total, we analysed 336 experiments, resulting in 404.880 trained neural network models (including the re-trained models for each saliency method). The number of experiments, result from the 21 datasets settings, the 8 models (4 different configurations $\times$ 2 model types) and 2 different types of train/test splits. We use an 8/2 split, for the 3inBinary and 2inQuarternary or 9/1 split for the rest, due to small dataset sizes. To further show some generalizability similar to~\cite{Schwenke2024SaliencyMA}, we further introduce datasets that utilize all data as training, validation and test set, thus making sure that the model has seen and potentially understood every input (Split Test vs Not Split Test). To make sure the model has learned the task without a bias, we primarily consider results where the base model reached 100\% acc. on the split test. Models with lower acc. as well as the Not Split Test case are only included in the Appendix \ref{ap:appendix} results to strengthen the generalizability of our conclusions, \cf the similarities between both cases in~\cite{Schwenke2024SaliencyMA}.

For training, the training data is always class balanced. The hyperparameters are selected over sample based manual optimization per dataset. For details about the pipeline, we refer towards our code\footnote{https://github.com/lschwenke/SaliencyMapsAmbiguous}.

\section{Base Model Results}\label{sec:results}
\begin{figure*}[htb!]
	\centering
	\includegraphics[width=1.9\columnwidth]{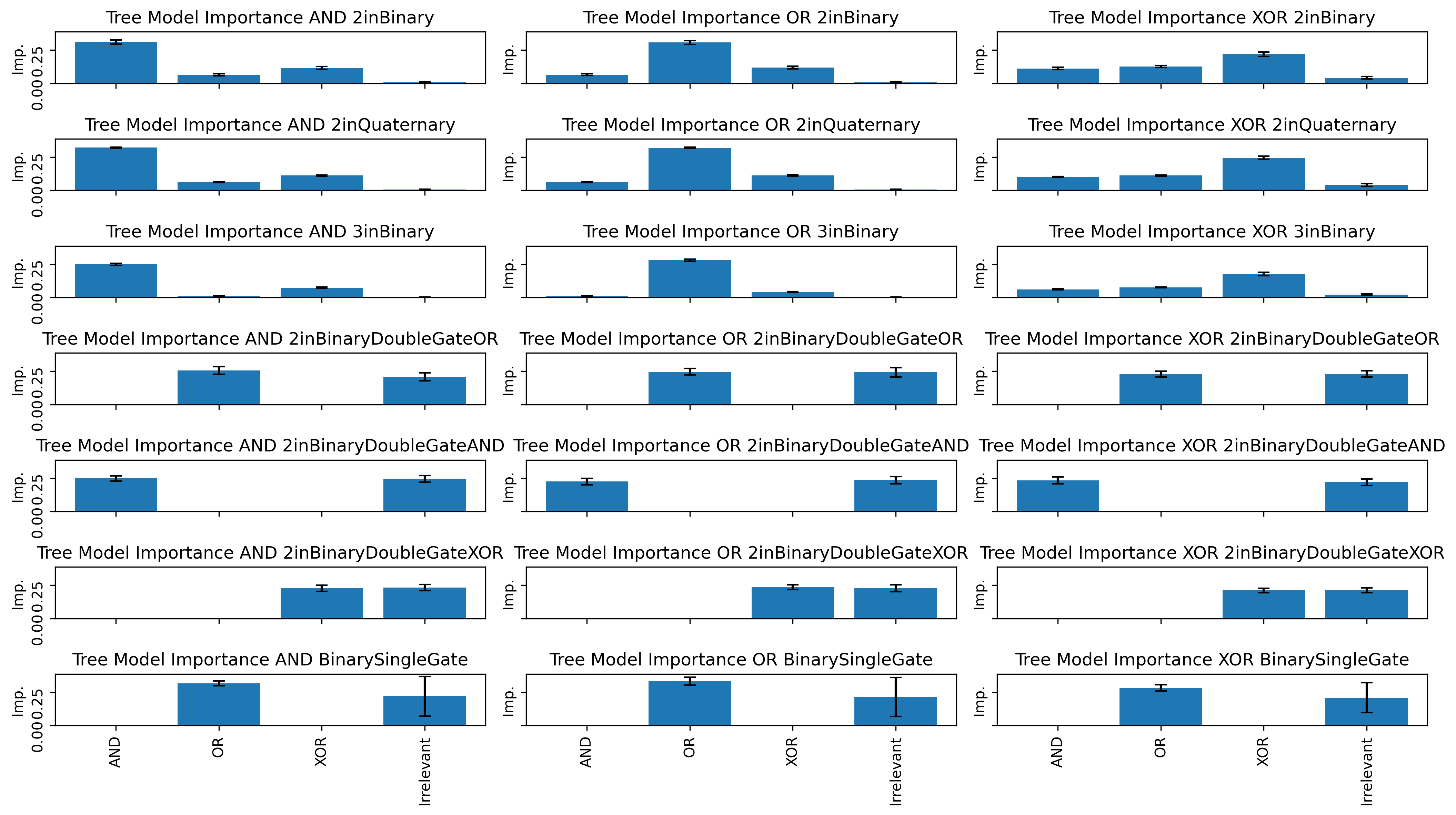}
	\caption{Average Tree Importance scores for the Random Forest models, differentiating between datasets and gates.}
	\label{fig:treeImps}
\end{figure*}

For our experiments around 75\% of the base models achieved an accuracy of 100\%. The Random Forest models often performed similarly well. While the global Tree Importance scores the irrelevant part close to our expectations, looking at the Tree Importance score for the smaller datasets in Figure~\ref{fig:treeImps}, we observe a relatively high scored irrelevant part, \cf to~\cite{Schwenke2024SaliencyMA}. The origin of this irregularity is unknown.
Therefore, in principle, this could also be a limiting factor for the neural network model evaluation.

To better structure the results, we primarily focus on different logical settings and rank the methods based on those. Using this, a more detailed analysis is possible, to see in which settings the results are more or less trustworthy. We focus on the following categories:
\begin{itemize}
    \item AND: Datasets that only have \textit{AND}-gates
    \item OR: Datasets that only have \textit{OR}-gates
    \item XOR: Datasets that only have \textit{XOR}-gates
    \item Complementary: Only datasets classes that need complementary information, \ie positive \textit{AND} or negative \textit{OR} (only for NIB- and GIB-Balanced)
    \item Redundant: Only datasets classes that need redundant information, \ie positive \textit{OR} or negative \textit{AND} (only for NIB- and GIB-Balanced)
    \item AND-OR: Datasets that have \textit{AND}-gates and \textit{OR}-gates
    \item AND-XOR: Datasets that have \textit{AND}-gates and \textit{XOR}-gates
    \item OR-XOR: Datasets that have \textit{OR}-gates and \textit{XOR}-gates
    \item AND-OR-XOR: Datasets that use all three types of gates
\end{itemize}

\subsection{Score Interpretation}
\label{sec:interpretations}

When interpreting saliency scores, the question is if negative values actually include class contradictory information or class relevant information; \eg for Shapley values the score represents the change of the target variable~\cite{molnar2020interpretable} and for gradient-based methods the scores can indicate the direction the input needs to be change to modify the output ~\cite{molnar2020interpretable}. Further, some methods perform only well on one removal strategy, MoRF or LeRF~\cite{tomsett2020sanity}. Hence, the interpretation of negative values might vary per method, where the influence of negative scores is not well analysed. We therefore explore three different settings per saliency method on how to interpret scores below zero: 
\begin{enumerate}
    \item AsIs: The score is interpreted as it is, \ie a negative score is less relevant than a positive score.
    \item Cutoff: As it is only relevant which information is needed for the given class, all negative values are mapped to 0.
    \item Absolute: The absolute value of each score is taken, as negative scores can also indicate class differentiable information.
\end{enumerate}

With these modes, we enable different alternatives for the interpretation of the respective values. Here, preprocessing towards the desirable interpretation per method might be needed, as other work has already shown certain flaws for many methods, \cf \cite{ju2021logic}. As we do not want to switch the interpretation mode per metric, we base our decision on which interpretation mode per saliency methods performed best for the NIB metric (both variations result in the same mapping); as capturing relevancy information is one of the prime goal of saliency scores. This results in the settings of Table \ref{tab:interpretMethods}, which we from now on assume as selected per saliency method.

Interesting is that \eg FeaturePermutation and KernelSHAP are performing best using the Cutoff method. This could be the case as those methods work using the change in the model outputs (which are often not integers). As the model is primarily trained on positive examples, negative examples might not be punished explicit enough, and hence the model performs worse for negative relevancy approximations. In contrast SHAP-IQ uses a model based baseline and therefore maybe avoids this problem.

\begin{table}[h!]
\centering
\caption{List of the best performing interpretation mode per saliency methods for the NIB metric (both NIB variations result in the same conclusion).}
\label{tab:interpretMethods}
\begin{tabular}{l|c|c|c}
\toprule
\multicolumn{1}{c|}{\textbf{Method}} & \multicolumn{1}{c|}{\textbf{Interpretation Mode}} \\ \midrule
LRP-Full~\cite{bach2015pixel} & Cutoff\\
LRP-Rollout~\cite{abnar2020quantifying} &  AsIs\\
LRP-Transformer~\cite{chefer2021transformer}  & AsIs\\
LRP-Transformer CLS~\cite{chefer2021transformer} & AsIs\\
Attention~\cite{vaswani2017attention} & AsIs \\
IntegratedGradients~\cite{sundararajan2017axiomatic}  & Absolute\\
DeepLift~\cite{shrikumar2017learning} & Cutoff\\
Deconvolution~\cite{zeiler2014visualizing}&  Absolute\\
GradCam~\cite{selvaraju2017grad} & AsIs\\
GuidedGradCam~\cite{selvaraju2017grad} & Absolute\\
GradCam++~\cite{chattopadhay2018grad} & AsIs\\
KernelSHAP~\cite{lundberg2017unified} & Cutoff\\
FeaturePermutation~\cite{molnar2020interpretable} & Cutoff\\
SHAP-IQ~\cite{fumagalli2024shap} & AsIs\\
\bottomrule
\end{tabular}

\end{table}

\subsection{Needed Information and Global Information}

\begin{figure*}[htb!]
	\centering
	\includegraphics[width=2\columnwidth]{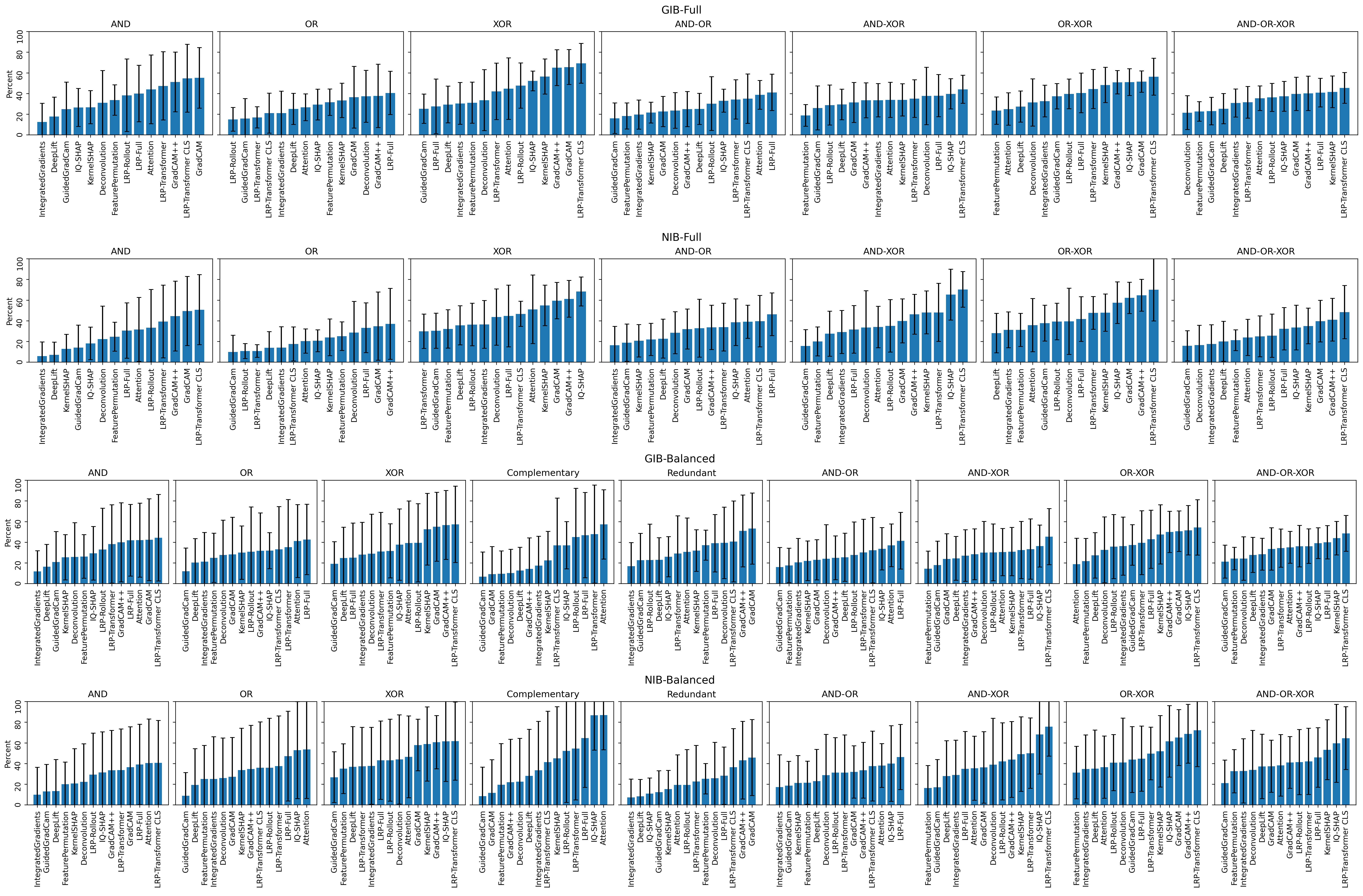}
	\caption{Average NIB/GIB-Full and NIB/GIB-Balanced results for each saliency method, ranked in and differentiated between the different scenarios. Only considering the split test data on 100\% acc. base models.}
	\label{fig:nib}
\end{figure*}

For the NIB and GIB metric, Figure \ref{fig:nib} shows the NIB/GIB-Full scores and the NIB/GIB-Balanced scores; only considering the split test data on 100\% acc. base models. As already discussed in~\cite{Schwenke2024SaliencyMA}, the less common classes per dataset tend to have a higher NIB, thus explaining the differences between the two metrics. This can also be seen between the \textit{Redundancy} and \textit{Complimentary} scenarios of the NIB-Balanced metric, showing that many methods struggle more with complimentary information, as redundant information is typically more present and thus simpler to capture. The question remains if the methods learn that certain classes are less common, even though the train data is balanced, or if the scores focus more about class differentiation, \ie a reasoning like: if it is not clearly class $c^1$ it is class $c^2$. While we only have two classes, it is neither the less a plausible scenario that can occur in real world data, sometimes even as a sub-scenario, where a complex decision can be reduced for specific instances to binary decisions. A few methods, like \eg especially GradCam and GradCAM++, rather struggle with redundant information, showing a different effect. Considering the GIB-Balanced, this difference between redundant and complementary information is less prominent, even though it is easier to miss redundant information.

While certain methods clearly perform better than others, no method could consistently capture all information in a desired manner (neither NIB nor GIB), defined by our assumptions \ref{sec:assumptions}. This means that while some methods capture specific scenarios better, in no analysed scenario any method provides fully trustworthy scores. 
Looking at the settings in Figure \ref{fig:nib} that include \textit{XOR}-gates, a general struggle of all methods can be seen. Notable, however, is that for the NIB-Full the scores are overall smaller for more complex scenarios, especially the \textit{AND-OR-XOR} scenario.
Further, notable for the results, are the similar standard deviation ranges for all scenarios (especially for NIB-Full), maybe indicating a property of the model.

Differentiating between both NIBs and GIBs, the ranking only shifts slightly. In a later section, we will have a closer look at the rankings for each metric to make conclusions for each scenario, and thus will first discuss primarily the general results. 
The results for the NIBs and GIBs on the not split any acc. setting in Appendix Figure \ref{fig:nibFull}, are for most cases similar, or somewhat bigger, highlighting the comparability of both settings.

\subsection{Logical Accuracy}

\begin{figure*}[htb!]
	\centering
	\includegraphics[width=2\columnwidth]{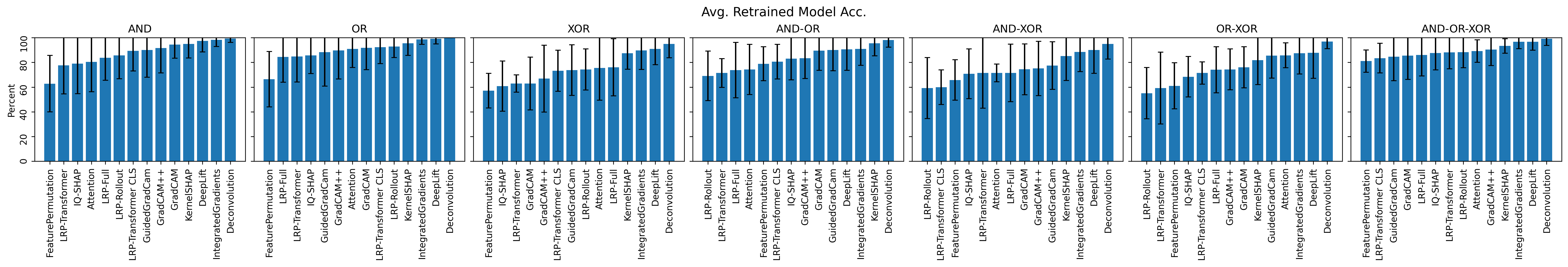}
	\caption{Average re-train performance after masking all data below or equal to the highest baseline input score per sample per saliency method, ranked in and differentiated between the different scenarios. Only considering the split test data on 100\% acc. base models.}
	\label{fig:lasaAcc}
\end{figure*}

\begin{figure*}[htb!]
	\centering
	\includegraphics[width=2\columnwidth]{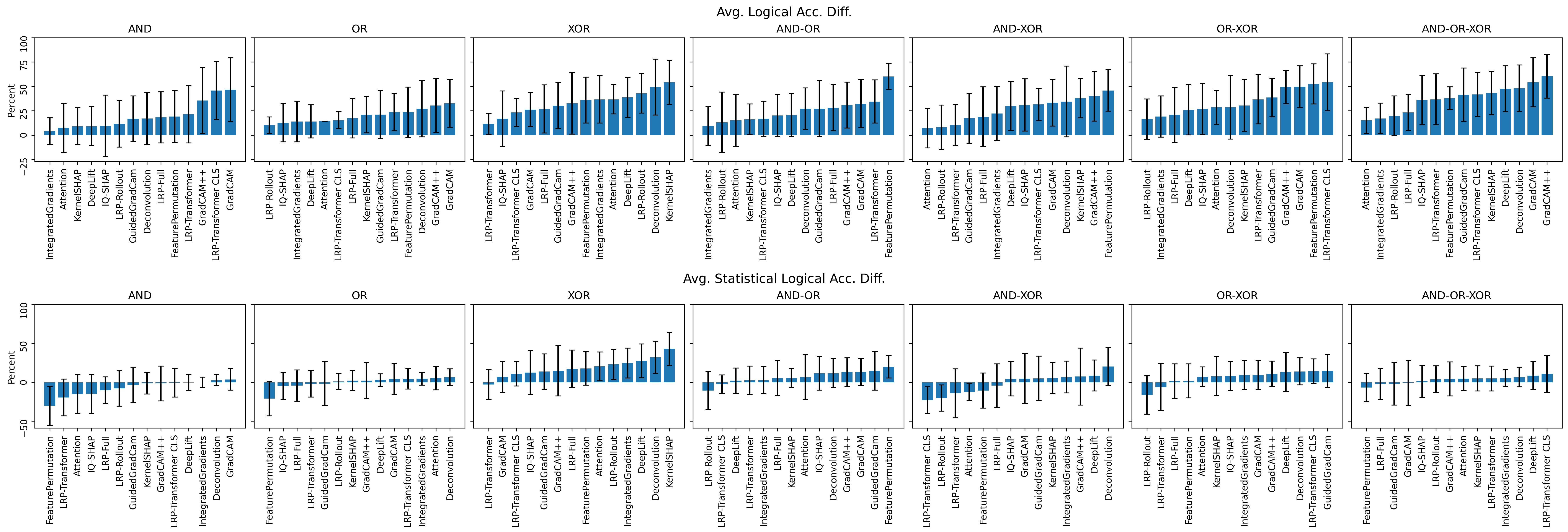}
	\caption{Average logical and statistical logical accuracy difference to the re-trained models per saliency method, considering the masked data and ranked per scenario. Only considering the split test data on 100\% acc. base models.}
	\label{fig:logicAccDiff}
\end{figure*}

The re-trained accuracy -- when taking the highest input baseline score per sample as a threshold -- can be seen in Figure~\ref{fig:lasaAcc}, only considering the split test data on 100\% acc. base models. While some methods like FeaturePermutation struggle to maintain accuracy, others perform to some extent quite well, sometimes methods even maintain the 100\% accuracy. However, as already shown in~\cite{Schwenke2024SaliencyMA}, many methods encode information into the masks as the logical accuracy fail to match the model performance. Figure \ref{fig:logicAccDiff}, shows the logical accuracy difference to the re-trained performance, to approximate how much information is extracted from the mask, only considering the split test data on 100\% acc. base models. We note, that even for simpler datasets assumption F is often not given. Again most methods could handle simpler datasets better, and again \textit{XOR} gates were a bit more challenging. Considering the logical statistical accuracy, the difference to the re-trained model performance in Figure \ref{fig:logicAccDiff} shows, that even statistical guessing cannot explain the performance for more complex datasets. While for the simpler models the statistical logical accuracy could match the model performance, the reason is likely different, as the training set was balanced. This gets further highlighted by the DCA results.

Appendix Figures \ref{fig:lasaAccFull} and \ref{fig:logicDiffFull} show the similarity of the results when considering all of our trained base models.

\subsection{DCA Results}

\begin{figure*}[htb!]
	\centering
	\includegraphics[width=2\columnwidth]{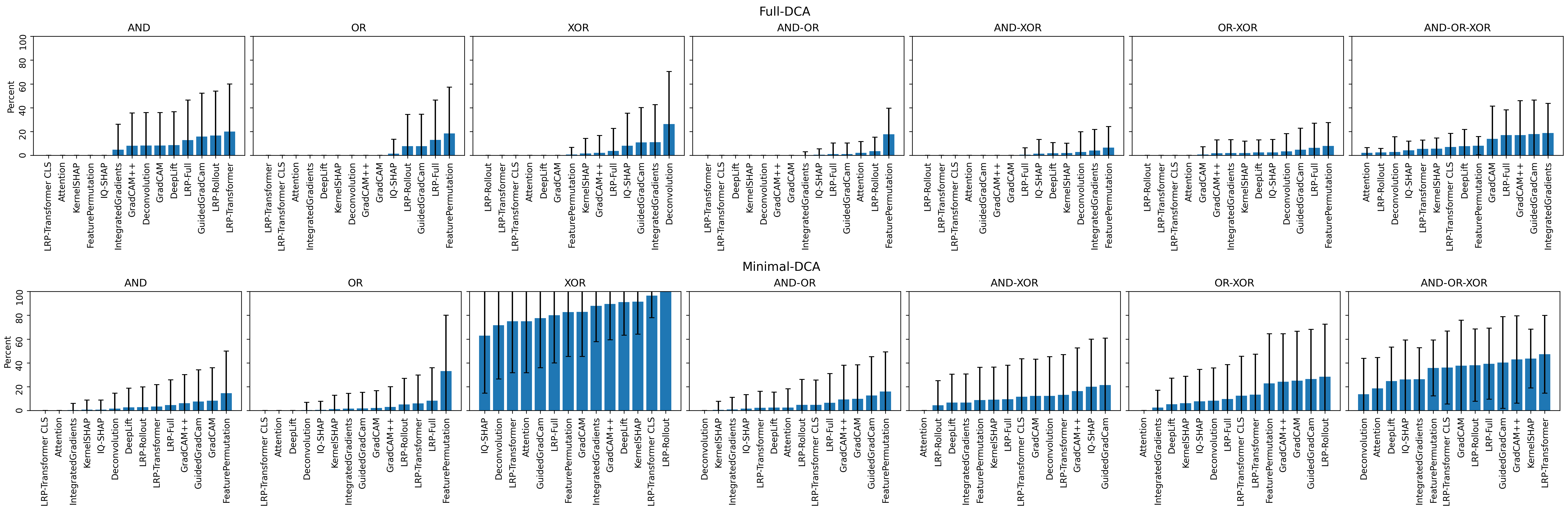}
	\caption{Average Full-DCA and minimal-DCA results for each saliency method, ranked in and differentiated between the different scenarios. Only considering the split test data on 100\% acc. base models.}
	\label{fig:dca}
\end{figure*}

If the logical performance is worse than the re-trained model, the question arises, where this information is encoded in. For this, we introduced the DCA metrics, covering two possible cases. The Full-DCA checks if information is leaked towards the always irrelevant inputs, hence leading to an encoding. Figure \ref{fig:dca} shows that for the Full-DCA in many scenarios, information is not leaked for many methods if the model fully understands the task at hand (split test set and acc. = 100\%); surprisingly, even for the XOR scenarios. However, for the \textit{AND-OR-XOR} scenario, it can happen for all methods. As for no other scenario this problem occurred so consistently and the \textit{AND-OR-XO} scenario datasets are all larger, we assume that the dataset size could be a crucial factor for this problem. The rather high standard deviation also indicates that this problem is unstable and could be due to internal model differences that do not influence the final model accuracy. \citep{shah2021input} also argues that this leakage of information occurs due to not robust enough models, \ie they showed that with robust training the model can be stabilized. Nonetheless, explainability should work on all models, because robustness is hard to fully verify.

With the minimal DCA, we check if information is leaked towards inputs that could include information, but should not in this particular input combinations. In Figure \ref{fig:dca} rarely cases exist with no leakage, and especially \textit{XOR} seems to be a problem. However, in more mixed scenarios the influence of \textit{XOR} reduces, compared to the \textit{XOR} scenario. With bigger and more complex scenarios, the occurrence increases, \cf the \textit{AND-OR-XOR} scenario. Here as well, a bigger dataset could have influence on the leakage, but our results that the complexity does so as well. The high standard deviation indicates again the model dependence of the problem.

As we noted in~\cite{Schwenke2024SaliencyMA}, those forms of leakage that leads to information encoding are only some possible cases. For example, the attention method had a DCA of zero for many scenarios, but still has a discrepancy in logical accuracy, indicating an encoding in different forms. In general, fixing information leakage though masks is hard, \cf~\cite{sturmfels2020visualizing, rong2022consistent}. A reason for the XOR encoding could be complexity, as XOR isn't linearly approximable. Hence, when aggregating the saliency scores into 1D space, problems occur. As such problems even occur on mathematically founded models like SHAP, multiple higher order SHAP methods are being proposed~\cite{fumagalli2024shap, harris2021joint, tsai2023faith, zhang2021interpreting, tsang2020does, janizek2021explaining}.

Considering Appendix Figure \ref{fig:dcaFull}, a small but overall higher error can be seen for most methods. Especially the higher DCA-Full error is interesting, showing that the model might put more relevancy towards irrelevant inputs. If this is due to the bigger dataset sizes or suboptimal models is unclear.

\subsection{Correlation}

\begin{figure*}[htb!]
	\centering
	\includegraphics[width=2\columnwidth]{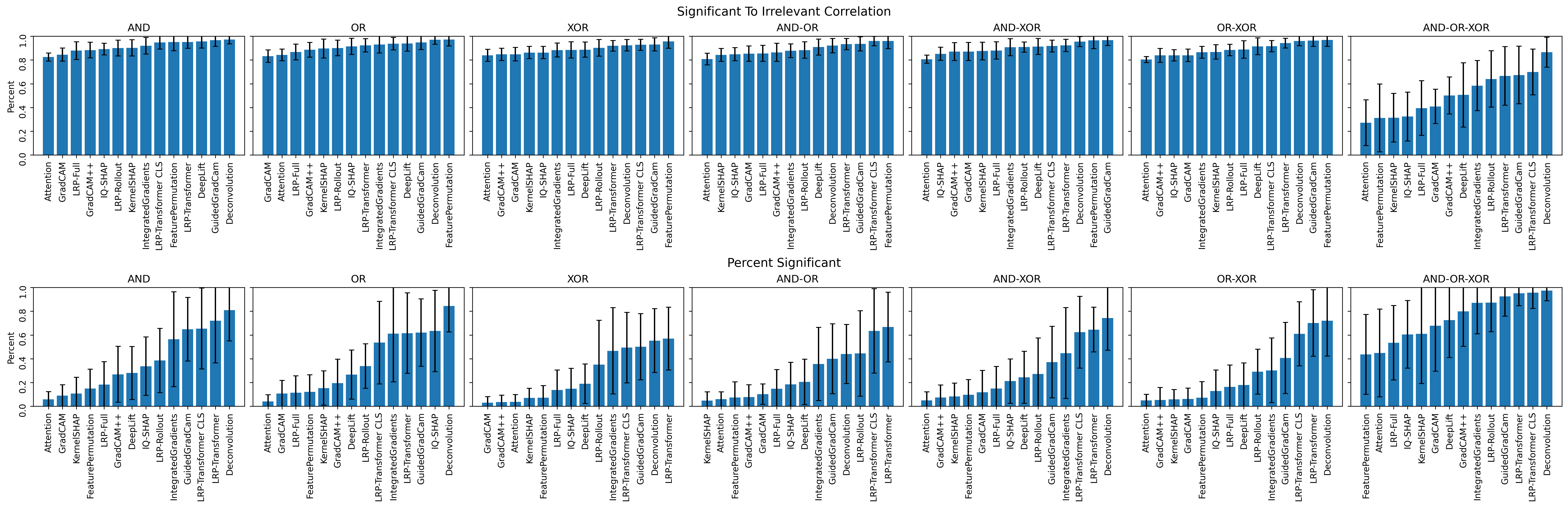}
	\caption{Average significant (p $<$ 0.5) Pearson correlation from the baseline scores to all other scores from the sample, averaged over all models (top) and average (over all split test data on 100\% acc. base models) percentage of correlations that are significant from all possible ones (bottom). Results ranked in and differentiated between the different scenarios.}
	\label{fig:correlation}
\end{figure*}

Because the irrelevant inputs (baseline) are not correlated towards the output class, relevant inputs should not correlate towards the irrelevant inputs as well (Assumption I). Figure \ref{fig:correlation} (top) shows the average significant (p $<$ 0.5) Pearson correlation between all gate inputs towards the irrelevant inputs, averaged over all base models (regardless of split and performance). We differentiate between the significant and not significant correlations, as Figure \ref{fig:correlation} (bottom) shows how many of the possible correlations are significant per method and scenario. In most scenarios, the correlation is really high, but the number of significant cases quite low. As the number of samples for the simpler scenarios is also rather low, this makes this data only partially meaningful. However, the \textit{AND-OR-XOR} scenario has a higher sample count and thus more significant cases, hence a better approximation of the correlation is given. Considering the ranking of the other scenarios, the ranking is still quite similar. Overall the correlations are still rather high, for actually uncorrelated data, hence strengths the assumption that the complex model internal representation of the problem is not perfect, but well enough to reach 100\% acc. This means that the numeric real valued output is complex and can only be partially used for relevance approximation, \ie changing irrelevant inputs will likely (slightly) change the numeric output. While actual linear problems like \textit{AND} and \textit{OR} are solvable using linear explanations like SHAP, LIME, etc., model outputs for linear problems are not linear. This results in an aggregation of problems, as model agnostic methods still often failed to capture the NIB, even though the model works. This is also somewhat represented in the results, as nearly all backpropagation-based methods have a high correlation (working with the complex model interia the most); in the middle of the ranking are the CAM based methods (working with one internal state); and perturbation-based methods have the lowest correlation (only looking at an aggregated output). Attention is here a special case, being in nearly all scenarios the method with the lowest correlation, while also have very few significant correlation overall.

To further analyse this problem, models with 0 error (in train, validation and test) are needed; which, however, are very hard to create. While this error should be considered, because the model still concludes the correct predictions per sample, interpretability methods are still expected to approximate good relevancy scores, making our assumptions still valid.   
As the meaning and expected results of this correlation are hard to evaluate -- as it might actually represent model internal processes -- we exclude this metric from the overall ranking.

As the not split setting has more data, the correlations in Appendix Figure~\ref{fig:correlationFull} are more refined as well and have more significant correlations. Comparing the data shows a consistency between the results.

\subsection{GCR Results}
\label{sec:gcrResults}

\begin{figure*}[htb!]
	\centering
	\includegraphics[width=2\columnwidth]{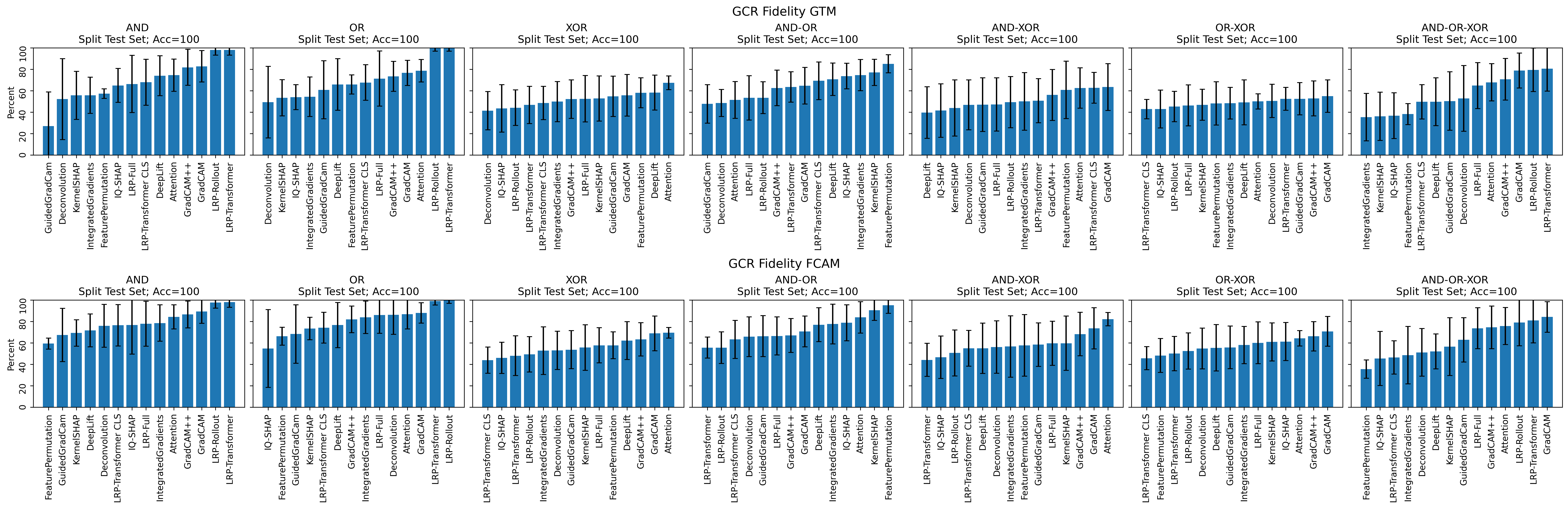}
	\caption{Average GCR fidelity results for each saliency method, ranked in and differentiated between the different scenarios. Only considering the split test data on 100\% acc. base models.}
	\label{fig:gcracc}
\end{figure*}

For the GCR we look at the relative average based representations (the average value an input value $\times$ input position receives). Using the GCR we want to use the globally averaged saliency scores as weights to enable relevancy based classification. Figure \ref{fig:gcracc} (top) shows the GTM option fidelity results and Figure \ref{fig:gcracc} (bottom) the fidelity results for the FCAM. They show that the performance for certain saliency methods is better than expected when considering the NIB and GIB (Figure \ref{fig:nib}). This is also due to the global input dependant view the GCR provides, which is about differentiability; \ie if all possible inputs for a specific input position are equally high scored, all inputs are also equally unimportant. Hence, the GCR puts every saliency score into the context of all alternative inputs that could occur, enabling a new interpretation. While this interpretation of saliency values breaks Assumption A, it shows a more global differentiation of inputs between samples. This is especially true for the FCAM, which shows the benefits of 2nd-order saliency scores, even if they were just up-scaled. Noticeable is that model internal representation (Attention- or CAM-based) often perform very good in differentiability. In some cases, even when up-scaled (\cf GradCAM). An exception for this the \textit{AND-OR} scenario, where perturbation-based methods like FeaturePermutation, KernelSHAP and IQ-SHAP perform especially well, while in many other scenarios they perform rather bad. We think this has to do with the assumptions made in these types of methods, which fit especially well for this scenario; as \cf~\cite{ju2021logic} already argued, needs to be considered and can often happen in evaluation. Considering more complex domains like \eg images and text, we think that in many tasks an \textit{AND-OR}-like relation is between the classes; as the classes are made out of specific patterns than can occur at different positions. Rarely, a specific input explicitly excludes other inputs. This get further highlighted, due to the sparsity of input combinations that have actual defined classes, \eg what to expect from a totally white input?
For this reason, for many tasks FeaturePermutation and SHAP-based methods might be a good explanation approach, hence explaining the success of said methods.

On the other hand, methods that perform well in local information capturing (NIB/GIB), do not necessary perform similar well globally, \eg IntegratedGradients. This means that our assumption I probably is not given for those methods, \ie a score of 5 in one sample could have a different meaning in another sample.

Considering the results in Figure \ref{fig:gcracc}, a general challenge is again the XOR gates. While it is possible to model XOR with the FCAM, it is rather tricky; as often all inputs are needed to derive XOR. Hence, for a simple explanation $n$-order scores are needed to explain an $n$ input XOR.

\subsection{Threshold GCR Results}

\begin{figure*}[htb!]
	\centering
	\includegraphics[width=2\columnwidth]{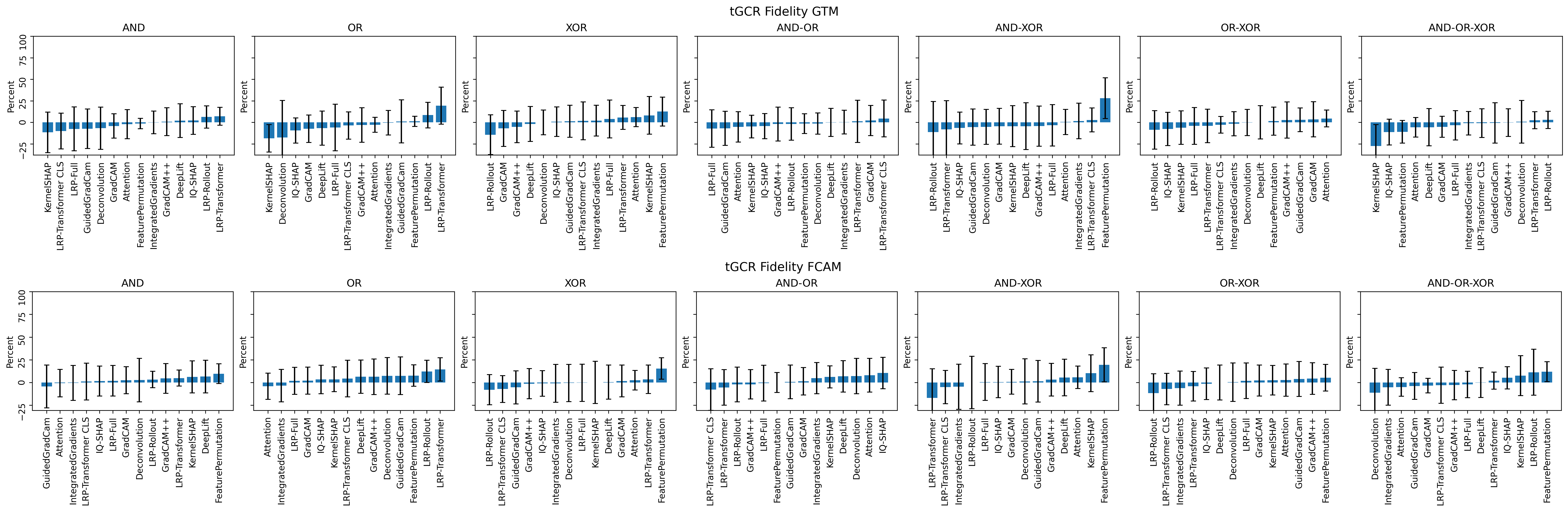}
	\caption{Average tGCR fidelity results for each saliency method, ranked in and differentiated between the different scenarios. Only considering the split test data on 100\% acc. base models.}
	\label{fig:tgcr}
\end{figure*}

Because this differentiability could also occur due to the encoding, Figure \ref{fig:tgcr} shows the average difference of the tGCR fidelity to the GCR fidelity, \ie the GCR fidelity change when ignoring all values that are equal or smaller to the highest baseline input score per sample. While a few methods show decreased performance, some methods often maintain or even increase their performance without the noise from the baseline scores. Nonetheless, for certain methods this further shows that at least some differentiability is included in the baseline (\cf Figure \ref{fig:gcracc}), as the baseline inputs created at least some part of this differenciation. 
Overall, for many methods the GCR still enables a new refined view on the saliency score, based on differentiation. In Appendix Figure \ref{fig:gcrFull} and \ref{fig:tgcrFull} the GCR and tGCR results are shown over all trained models; showing a very close similarity to the split test 100\% accuracy results.

\begin{table*}[!htb]
\scriptsize
\centering
\caption{Average scores per metrics $\times$ saliency methods over all datasets, only considering the split test data and models reaching an accuracy of 100\%.}
\label{tab:avgScore}

\begin{tabular}{@{}lrrrrrrrrrrrrrr@{}}

\toprule \vspace{1.7cm}\\
                         & \myrotcell{LRP-Full} & \myrotcell{LRP-Rollout} & \myrotcell{LRP-Transformer} & \myrotcell{LRP-Transformer CLS} & \myrotcell{Attention} & \myrotcell{IntegratedGradients} &\myrotcell{ DeepLift} & \myrotcell{KernelSHAP} & \myrotcell{GuidedGradCam} & \myrotcell{FeaturePermutation} & \myrotcell{Deconvolution} & \myrotcell{GradCAM++} & \myrotcell{GradCAM} & \myrotcell{IQ-SHAP} \\ \midrule
NIB Balanced                  & 46.03    & 34.80       & 35.92           & 53.00               & 48.07     & \textbf{26.37}             & 32.22    & 45.16      & 27.62         & 35.14              & 31.93         & 50.36     & 46.48   & 51.92   \\
GIB Balanced                  & 39.25    & 29.31       & 29.46           & 43.86               & 42.55     & \textbf{21.52}              & 26.10    & 35.79      & 22.11         & 27.61              & 25.09         & 42.32     & 38.94   & 35.38   \\
NIB Full                      & 38.94    & 28.91       & 32.40           & 51.94               & 35.05     &\textbf{24.55}               & 27.27    & 40.44      & 26.77         & 31.55              & 31.00         & 52.57     & 49.39   & 44.83   \\
GIB Full                      & 37.54    & 29.15       & 31.19           & 46.41               & 35.59     & 24.65               & 27.85    & 38.03      & \textbf{24.16}        & 28.03              & 27.34         & 46.01     & 44.12   & 37.48   \\
Logical Acc. Diff.            & 32.44    & 27.02       & 36.61           & 42.71               & 28.78     & \textbf{22.34}               & 35.97    & 38.72      & 41.21         & 40.38              & 34.03         & 53.80     & 52.23   & 37.44   \\
Statistical Logical Acc. Diff & 10.82    & 12.55       & 14.42           & 15.24               & 13.02     & 13.95               & 16.85    & 16.72      & 14.79         & \textbf{3.52}               & 16.55         & 20.38     & 20.75   & 20.00   \\
Full DCA                      & 31.07    & 14.92       & 23.71           & 17.71               & 11.51     & 11.19               & 10.97    & \textbf{6.40}       & 23.53         & 14.29              & 7.69          & 19.75     & 22.04   & 12.01   \\
Minimal DCA                   & 43.05    & 36.18       & 48.20           & 29.71               & 20.76     & 22.06               & 24.66    & 31.35      & 42.33         & 46.09              & \textbf{20.25}         & 40.17     & 39.33   & 27.44   \\
GCR FCAM Acc.                 & 66.17    & 70.48       & 69.24           & 60.35               & \textbf{76.48}     & 63.36               & 62.25    & 67.00      & 58.66         & 64.04              & 58.06         & 67.28     & 74.62   & 60.53   \\
GCR GTM Acc.                  & 55.70    & 68.37       & \textbf{69.17}           & 57.93               & 60.02     & 50.07               & 57.97    & 51.09      & 49.50         & 58.21              & 50.01         & 61.39     & 66.96   & 52.89   \\ \bottomrule
\end{tabular}
\end{table*}

\subsection{Result Ranking}
\label{sec:ranking}

To enable a summarized view on which method performs in which scenarios how well on which metric, we introduce two different rankings; each based on the split data 100\% acc. models. Because multiple metrics measure similar properties, we categorize them into four different properties:
\begin{itemize}
    \item Information capturing: The ability to rank local inputs correctly, based on relevant and not relevant inputs, \ie the two versions of NIB and GIB. A higher score means, more mistakes in capturing information.
    \item Truthfulness of classification: How well does the class discriminatively after masking the data actually represent class relevant information, \ie  Logical Acc. difference and Statistical Logical Acc. difference. A lower score means that the found differentiation of relevant information is actually in the numeric score values included and not encoded in any other way.
    \item Information leakage: How strongly does the method tend to encode/leak information into not relevant inputs if relevant information is lost, \ie Full-DCA and Minimal-DCA. A higher score means that information is leaked more often. While this has a connection to the truthfulness of classification, as this also come from encodings, the DCA encoding are rather specific places information leaks to. Our results show for many methods a deviation in rank.
    \item Global differentiability: How strongly can the scores be class discriminative in a global context, \ie the relative average based GTM and FCAM. A higher score means that more information can be extracted out of the scores when using them as weights.
\end{itemize}

\begin{table*}[!htb]
\scriptsize
\centering
\caption{Average rank per metric properties $\times$ saliency methods, using the average scores per metric as ranking basis from Table \ref{tab:avgScore}.}
\label{tab:avgScoreRank}
\begin{tabular}{@{}lrrrrrrrrrrrrrr@{}}
\toprule \vspace{1.7cm}\\
                         & \myrotcell{LRP-Full} & \myrotcell{LRP-Rollout} & \myrotcell{LRP-Transformer} & \myrotcell{LRP-Transformer CLS} & \myrotcell{Attention} & \myrotcell{IntegratedGradients} &\myrotcell{ DeepLift} & \myrotcell{KernelSHAP} & \myrotcell{GuidedGradCam} & \myrotcell{FeaturePermutation} & \myrotcell{Deconvolution} & \myrotcell{GradCAM++} & \myrotcell{GradCAM} & \myrotcell{IQ-SHAP} \\ \midrule
Information capturing      & 9.75     & 5.25        & 7.00            & 13.75               & 10.00     & \textbf{1.25}                & 3.75     & 9.50       & 1.75          & 5.50               & 3.50          & 12.75     & 11.00   & 10.25   \\
Truthfulness of classification  & 3.00     & \textbf{2.50 }       & 6.50            & 10.00               & 3.50      & 3.00                & 8.50     & 9.50       & 9.00          & 5.50               & 7.00          & 13.50     & 13.50   & 10.00   \\
Information Leakage & 13.00    & 8.00        & 13.50           & 7.50                & 3.50      & 3.50                & 3.50     & 4.00       & 11.50         & 10.00              & \textbf{1.50 }         & 10.00     & 10.00   & 5.50    \\
Global differentiability          & 8.00     & \textbf{2.50}        & \textbf{2.50}            & 10.00               & 3.00      & 10.50               & 8.50     & 8.50       & 13.50         & 7.00               & 13.50         & 4.50      & \textbf{2.50}   & 10.50   \\ \midrule
Avg. Rank                     & 8.44     & \textbf{4.56}        & 7.38            & 10.31               & 5.00      & \textbf{4.56}                & 6.06     & 7.88       & 8.94          & 7.00               & 6.38          & 10.19     & 9.25    & 9.06    \\ \midrule
Overall Ranking              & 9.00     & 1.00        & 7.00            & 14.00               & 3.00      & 1.00                & 4.00     & 8.00       & 10.00         & 6.00               & 5.00          & 13.00     & 12.00   & 11.00   \\ \bottomrule
\end{tabular}
\end{table*}

Table \ref{tab:avgScore} shows the averaged scores over all datasets per metric. We rank all the methods and show the average ranking per property group in Table \ref{tab:avgScoreRank}, showing, regardless of the scenario, which methods perform well for which metric. In most cases the methods rank similar per metric which belongs into the same group. This results in IntegratedGradients, LRP-Rollout and Attention being the best overall metrics. IntegratedGradients struggle with the differentiability, while LRP-Rollout has a information leakage problem and Attention has bad information capturing. For Attention, this could mean that the score representations just need to be fine-tuned, as the differentiability is quite high. As LRP-Rollout is a specific aggregation of Attention, it seems to solve a few problems, but due to suboptimal aggregation information is leaked. As IntegratedGradients captures information better than any other method, the differentiability probably suffers due to not globally comparable scores.

A different view based on scenarios is provided in Table \ref{tab:avgRankPerScenario}, showing the average ranking per logical scenario per metric. The average ranking is based on the average scores per grouped property, which is based on the average ranking per metric. Because different metrics cannot be aggregated, the average ranking is considered, rather than average scores. Interesting is that the ranking over the average of all scenarios is similar to the ranking from Table \ref{tab:avgScoreRank}, even though the ranking does not represent numeric differences; \eg just because one method is ranked higher than another, the actual score difference is unclear. In general, Table \ref{tab:avgRankPerScenario} shows how well different methods can be used for different scenarios. Overall this is still an approximation and for specific conclusions, the specific data from previous sections should be considered.

\begin{table*}[!htb]
\scriptsize
\centering
\caption{Showing the average rank over all metrics for each scenarios $\times$ saliency methods, only considering the split test data and models reaching an accuracy of 100\%. The average ranking is based on the average scores per grouped property.}
\label{tab:avgRankPerScenario}
\begin{tabular}{@{}lrrrrrrrrrrrrrr@{}}
\toprule \vspace{1.7cm}\\
                         & \myrotcell{LRP-Full} & \myrotcell{LRP-Rollout} & \myrotcell{LRP-Transformer} & \myrotcell{LRP-Transformer CLS} & \myrotcell{Attention} & \myrotcell{IntegratedGradients} &\myrotcell{ DeepLift} & \myrotcell{KernelSHAP} & \myrotcell{GuidedGradCam} & \myrotcell{FeaturePermutation} & \myrotcell{Deconvolution} & \myrotcell{GradCAM++} & \myrotcell{GradCAM} & \myrotcell{IQ-SHAP} \\ \midrule
AND             & 9.06          & 6.06          & 6.75            & 10.88               & 6.63          & \textbf{4.50}                & 5.88          & 7.69          & 7.44          & 7.81               & 8.13          & 8.88          & 9.56          & 5.75          \\
OR              & 9.13          & \textbf{3.75}          & 6.38            & 7.06                & 5.38          & 5.06                & 7.75          & 9.31          & 7.31          & 8.94               & 7.38          & 9.88          & 9.50          & 8.06          \\
XOR             & 8.06          & 8.56          & 8.19            & 7.38                & 7.19          & 4.75                & 5.38          & 8.75          & 9.25          & \textbf{3.50}               & 5.25          & 10.13         & 9.25          & 9.38          \\
AND-OR          & 11.75         & 6.94          & 10.06           & 7.44                & 7.75          & \textbf{2.44 }               & 5.25          & 3.94          & 9.13          & 7.81               & 5.38          & 10.00         & 8.94          & 8.19          \\
AND-XOR         & 8.38          & 6.38          & 8.19            & 8.00                & \textbf{3.81}          & 7.00                & 5.06          & 7.63          & 8.31          & 5.81               & 7.06          & 9.88          & 8.88          & 10.63         \\
OR-XOR          & 8.19          & 5.19          & 8.44            & 10.50               & \textbf{5.13}          & 5.81                & 5.69          & 5.75          & 8.50          & 6.44               & 6.50          & 10.13         & 9.75          & 8.88          \\
ALL             & 7.31          & 5.63          & 6.56            & 11.06               & \textbf{3.38}          & 7.13                & 7.56          & 10.25         & 6.69          & 7.00               & 6.50          & 9.75          & 7.88          & 8.31          \\ \midrule
Avg. Rank       & 8.84     & 6.07        & 7.79            & 8.90                & 5.61      & \textbf{5.24}                & 6.08     & 7.62       & 8.09          & 6.76               & 6.60          & 9.80      & 9.11    & 8.46    \\ \midrule
Overall Ranking & 11.00         & 3.00          & 8.00            & 12.00               & 2.00          & 1.00                & 4.00          & 7.00          & 9.00          & 6.00               & 5.00          & 14.00         & 13.00         & 10.00         \\ \bottomrule
\end{tabular}
\end{table*}

\section{Discussion}
\label{sec:discussion}

With our analysis, we showed that all attribution based explanation methods fail to capture typical assumptions, \cf Section \ref{sec:assumptions}. While certain expectations work better for simple scenarios, no method could consistently capture all relevant information -- compared to our introduced irrelevant input baseline. We ranked the data to enable a deeper understanding in which scenarios for which property, which method performs best and hence is the most reliable or needs the most improvement. In the following, we further want to discuss conclusions and reasons for our results, as well as limitations of our experimentation.

\subsection{Result Conclusions and Assumptions}

One typical challenge for many properties was the \textit{XOR}-gate. As already shown by other work, even mathematically funded methods like SHAP struggle with nonlinear problems, as they can aggregate information in a way that can be misleading~\cite{konig2024disentangling}. Hence, one problem of current explanation methods is that they primarily work in the first order, \ie assign one relevancy value to one input, thus neglect dependant input interactions. For this reason, we argue for the use of higher order explainability, that considers interactions betweens inputs. Nonetheless, even in first order scores, higher order relations can be extracted, \cf FCAM in Section \ref{sec:gcrResults}. Especially for GradCAM this often worked very well.

 In Section \ref{sec:ranking} we analysed different logical scenarios and how well each method performs. How can those results be applied to different domains? Considering \eg image data: As the image input domain is quite big, object classes in images are all about specific combinations of inputs (AND) from a brought selection of possibilities (OR), with rarely very close classes (input vice). Additionally, inputs rarely exclude each other, \eg just because we see a dog this doesn't mean there is no cat. For \eg the text domain, many different inputs are far away (embedding vector distance), hence we argue it is again a lot about \textit{AND} and \textit{OR} relations, build from specific areas of words/meanings.
 Therefore, we argue that many domains are rarely having very complex logical interactions and thus the \textit{XOR} relation is rather an exception than a common occurrence. This gets further emphasized due to the large input domains and the not completely defined nature of those tasks, \eg what is expected from a complete 0 input. Hence, many classes are often rather quite far away. For this reason we think, that many interpretations techniques often work at least to some extent, as we have shown in the simple dataset scenarios. For example, localizing objects in images that represent the target object often works quite well, \eg~\cite{chefer2021transformer, lapuschkin2019unmasking}, hence at least to some extent shows relevant pixels. As discussed in Section \ref{sec:gcrResults}, this could also explain the popularity of FeaturePermutation and KernelSHAP, as many tasks would be similar to the \textit{AND-OR} scenario -- based on our argumentation.

One typical challenges in the interpretation of attribution based methods is to know what a specific value assignment means. In Section \ref{sec:interpretations}, we discussed that different approaches for numbers equal and below zero can be appropriate. Further, the GCR also shows that a globally relative interpretation can show up further interpretations of attribution scores; \ie considering global and local interpretation provides another challenges. Do the values are globally comparable, or is a local baseline needed to actually make any conclusions about input relations, as \eg we suspect for IntegratedGradients -- considering their information capturing abilities vs. their global differentiability. For example an interesting observation is that methods that use internal model states, \ie methods based on Attention (Attention, LRP-Rollout and LRP-Transformer) or CAM (GradCAM, GradCAM++) have a very good global differentiability, indicating that the internal numbers need some further processing to enable interpretability in a form as expecting from our assumption. In conclusion, for many methods, it is currently hard to tell how to interpret their numeric scores properly, as currently each method has undesired properties.
With our analysis, we hope to give more insight into this topic. We further argue that considering how relevancy methods are typically interpreted, a weight based interpretation for global conclusions that can fully ignore inputs is a desirable verification method. Here we suggest the GCR as a possible interpretation of this, but other methods can be considered as well.

FeaturePermutation and SHAP are model agnostic and mathematically founded. Considering the results on linear data from~\cite{tritscher2020evaluation}, methods like SHAP should be able to fully capture the linear synthetic datasets. However, in our experiments they do not. We highly suspect that this is due to model differences. As the model does not output the prediction as integer, but a real number: model agnostic methods need to understand a complex non-linear output linearly. The more complex the model, the more complex those changes can be. Hence, inputs that should be irrelevant for the output, can lead to changes in the output that could be aggregated intro scores that could suggest relevancy. Further considering that inputs cannot be ignored, a numeric baseline is needed for some approaches and thus introduces certain problems, \cf~\cite{haug2021baselines}.

\subsection{Limitations}
As discussed in the last section, models have real valued outputs: \ie even though we primarily considered methods that fully understood the task (100\% acc on all possible data samples), our results still include small errors, as the model output does not have an error of 0. Because neural networks have very complex decision paths, multiple factors, \eg irrelevant inputs, could have an unknown effect towards the results, as different inputs can open up different decision paths. Hence, we can only approximate the model's reasoning based on a minimal set of needed information in an as good as possible fashion, considering the task. For a better approximation, the error function would need to be 0, which is however hard to archive. As we focus on typical explanations and expectations, coping with these challenges is important, as well as analysing what to expect.
Anyhow, as our inputs are independent and a basic understanding of the task is needed to perform well on the test data, it is still expected that relevant inputs should be scored the highest, to actually represent relevant classification information. Hence, our experiments are still a good approximation towards our assumptions from Section \ref{sec:assumptions}.

A further limitation on our experiments is that our simple datasets are rather small -- compared to the 3 datasets from the \textit{AND-OR-XOR} scenarios -- and hence for certain scenarios quaternary inputs as missing. This could be a further interesting experiment case, to deepen the understanding and analysis of saliency methods.
As different methods work differently, the optimal aggregation from second order saliency scores to first order scores or the up-scaling from first to second order, can be different per method. While better methods might exist, we showed that our approaches at least perform to some extent quite well. Nonetheless, better options could exist.

While we found certain methods work better in specific scenarios, we cannot be 100\% sure they might perform similar on different tasks~\cite{kokhlikyan2021investigating, yona2021revisiting}, due to the complexity of each model. However, as we discussed before, on more complex datasets this approximation is not feasible. We still argue, that due to the basic relation \andor contains (redundant, complimentary, exclusive), similar cases can and will occur in more complex real world scenarios and hence scenarios will exist that fail to capture the minimal information coverage, as we have shown with our experiments. \Ie a good explanation should work in simple as well as in complex scenarios. As a simple scenario can easier occur, errors could even be further accumulated.

While for certain methods the GCRs performed very well and we argue towards a weight-based interpretation, the GCR is only one of many possible ways to do so. As~\cite{ju2021logic} discussed, the fundamental structures of specific evaluations could favour similar methods. Therefore, the GCR can also be interchanged with different methods. 

\subsection{Method Summary}

In the following, we provide a table with the relative strengths and weaknesses of each method, as a quick summary.

\begin{itemize}
    \item \textbf{LRP-Full}: The classic LRP variant is good in truthfulness, \ie the information the re-trained model finds somewhat close to the actual relevant information. But the overall information capturing is quite low, as well as the information leakage of lost information quite bad. Overall, the assumption made in LRP seem to be suboptimal, as the method is at the latter part of our ranking.
    \item \textbf{LRP-Rollout}: This specific aggregation of Attention matrices in the Transformer architecture performs very well in global differentiability and truthfulness, and is overall one of our top 3 methods. \Ie the numeric interpretation might need some further processing, but definitely contains class relevant information, as the information capturing abilities are mediocre, but the global differentiability is compared to all other methods the best. The Attention aggregation from LRP-Rollout, however, seems to introduce some information leakage, \cf the Attention method. Further interesting is that this method performed better than all other methods on \textit{OR} related data, \ie datasets with a lot of class positive redundant information.
    \item \textbf{LRP-Transformer}: This extension of LRP in combination with the Attention-Rollout is very globally differentiable, but fail in all other aspects. However, this could again mean that the numeric values just need some further processing to fulfil our assumptions. 
    \item \textbf{LRP-Transformer CLS}: This variation, using an addition input to capture first order relevancy, does not work well for our data; with no specific advantages.
    \item \textbf{Attention}: This very basic form of Attention aggregation performed well in truthfulness, information leakages and global differentiability, hence is one of the top 3 methods. Especially the performance as second order metrics was very good, \ie eventually the reduction into first order was suboptimal, hence the low performance in information capturing. Anyhow, as the global differentiability is very good, we suspect that another refining step could also improve the information capturing ability. Further to note is that Attention performed by far the best in the \textit{AND-OR-XOR} and \textit{AND-XOR} scenario, and the best for the \textit{OR-XOR} scenario, suggesting that specific complex situations are captured better than in other metrics; thus further suggesting good performance on complex real world data. 
    \item \textbf{IntegratedGradients}: The method with the best information capturing, that is overall in the top 3 ranking. Only the global differentiability is lacking, suggesting not globally comparable scores. Especially, the performance on the \textit{AND} and \textit{AND-OR} scenario was very good, hence suggesting a very good applicability on \eg image data.
    \item \textbf{DeepLift}: An overall good performing method, that only lacks in global differentiability (\cf IntegratedGraients) and has a suboptimal truthfulness. As our analysed leakage is rather low, the encoding of information must happen in another way than we analysed. 
    \item \textbf{KernelSHAP}: Trying to capture model predictions using LIME doesn't seem to work too well, as only the information leakage is low. We think this is due to the real valued numeric nature of the model output. However, the method performed quite well in the \textit{AND-OR} scenario, suggesting the applicability on many real work problems, \cf Section \ref{sec:discussion}. Further, it has the lowest Full-DCA, \ie never relevant information is rarely overestimated. 
    \item \textbf{GuidedGradCAM}: It has one of the best information capturing, but lacks in all other metrics, \ie the information that is not captured is encoded. This can make interpreting this method somewhat hard, \eg as information is often leaked to irrelevant inputs. Overall, this method had the best GIB-Full performance.
    \item \textbf{FeaturePermutation}: An overall ok method, that leaks information, primarily between possible relevant samples (minimal-DCA). We think, as inputs can sometimes replace each other in permutation, some inputs get over or underestimated. Interesting is also the rather low statistical difference, which could result from the statistical global approach that FeaturePermutation has. Additionally, over all metrics, it performed very good in the \textit{XOR} scenario, maybe also resulting from the permutation approach.
    \item \textbf{Deconvolution}: Performing quite well in information capturing and has the lowest DCA leakage, but still encodes information in another way; seen by the bad truthfulness. Also has as bad global comparability, \cf IntegratedGradients.
    \item \textbf{GradCAM++}: While having a good global differentiability, the other metrics are quite bad, hence maybe just need some further processing, \cf LRP-Transformer and Attention.
    \item \textbf{GradCAM} Similar to GradCAM++, but even though GradCAM++ should improve GradCAM, for our results the normal GradCAM performs slightly better. Especially, the global differentiability is really good, even when the values are upscaled to second order scores, indicating that information might be captured quite well, but the scores need some further interpretation processing.
    \item \textbf{IQ-SHAP}: Similar to KernelSHAP this method has a rather small information leakage. But overall it performs slightly worse than KernelSHAP. IQ-SHAP has a numeric baseline, which might introduce limitations due to the complex numeric outputs of the model.
    
\end{itemize}

\section{Conclusions}\label{sec:conclusions}
In this work, we extended our previous analysis on the logical dataset framework ANDOR, where we showed that all analysed saliency methods fail to grasp all needed classification information for all possible scenarios. We included more datasets and additional metrics, in order to better differentiate in which scenarios which saliency methods perform better. In addition, we also applied the Global Coherence Representation as an additional evaluation method in order to enable actual input omission and to evaluate higher order saliency scores.

Our results show that again, all methods fail to consistently capture all relevant information, regardless of the scenario. However, our analysis on different information leakages showed that certain methods rarely encode data for simple scenarios, \ie inputs that are found relevant, are actually relevant. But on more complex scenarios, the encoding effect increased, reducing the overall trust in all saliency methods. As each method has different fundamental assumptions, different methods perform in different scenarios differently well. In our overall analysis, IntegratedGradients, LRP-Rollout and Attention were the best performing methods.

While, in our results not all scores show the desired relevancy, our different interpretation modes and the GCR indicates that further refining the scores might help to find an underlying differentiability to improve the interpretability between classes. Overall, we argue here, that just assigning a value to an input is a bad form of explanation, which always needs contextual information to make sure the interpretation is correct. Hence, this highlights the need for improved methods, as well as interpretations for the interpretations.

For future work, we target exploring more stable training processes as well as evaluations on more complex real world data, in order to accordingly extend and exploit our evaluation methodology further. 

\bibliographystyle{ieeetr}
\bibliography{main.bib}

\clearpage
\onecolumn
\appendix
\label{ap:appendix}

\section{Appendix}
\label{app:appendix}
In the following, we provide addition plots, showing the results when considering all trained models, to provide further results supporting our discussion. Keep in mind that errors might be introduced due to suboptimal models.

\begin{figure}[ht!]
	\centering
	\includegraphics[width=0.99\columnwidth]{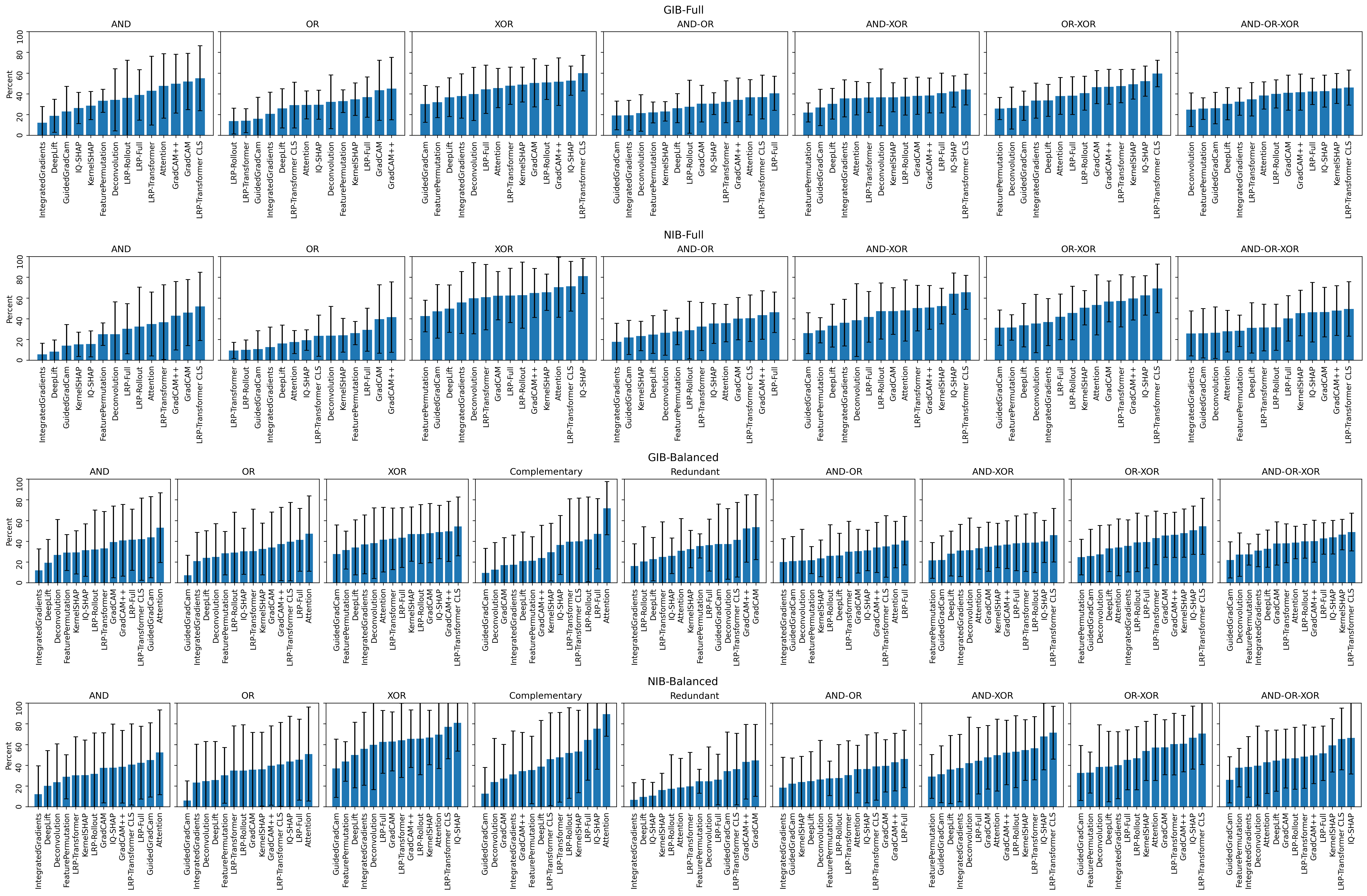}
	\caption{Average NIB/GIB-Full and NIB/GIB-Balanced results for each saliency method, ranked in and differentiated between the different scenarios. Considering all trained base models.}
	\label{fig:nibFull}
\end{figure}

\begin{figure}[ht!]
	\centering
	\includegraphics[width=0.99\columnwidth]{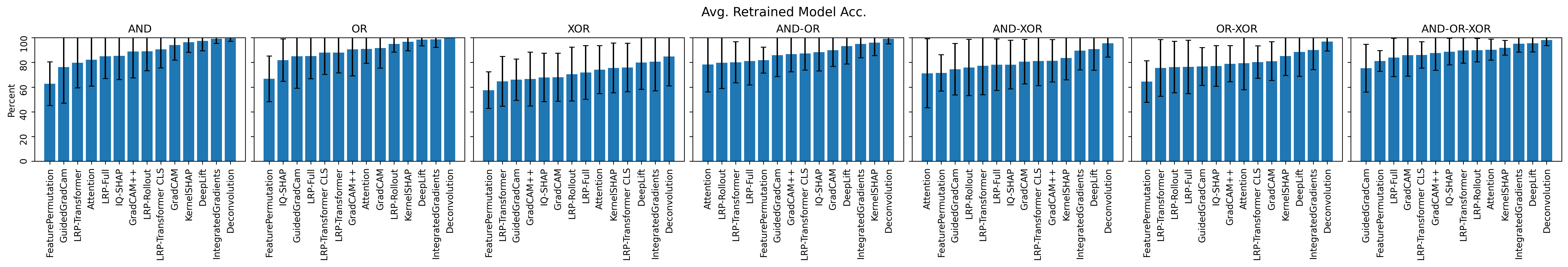}
	\caption{Average re-train performance after masking all data below or equal to the highest baseline input score per sample per saliency method, ranked in and
differentiated between the different scenarios. Considering all trained base models.}
	\label{fig:lasaAccFull}
\end{figure}

\begin{figure}[ht!]
	\centering
	\includegraphics[width=0.99\columnwidth]{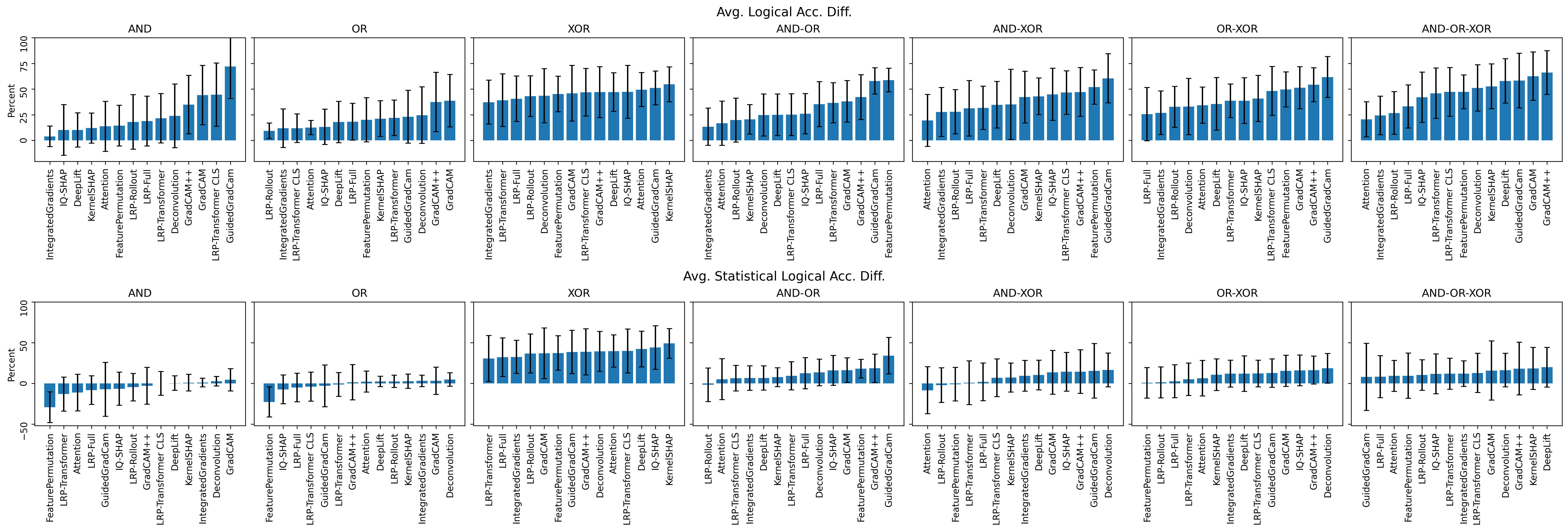}
	\caption{Average logical and statistical logical accuracy difference to the re-trained models per saliency method, considering the masked data and ranked per
scenario. Considering all trained base models.}
	\label{fig:logicDiffFull}
\end{figure}

\begin{figure}[ht!]
	\centering
	\includegraphics[width=0.99\columnwidth]{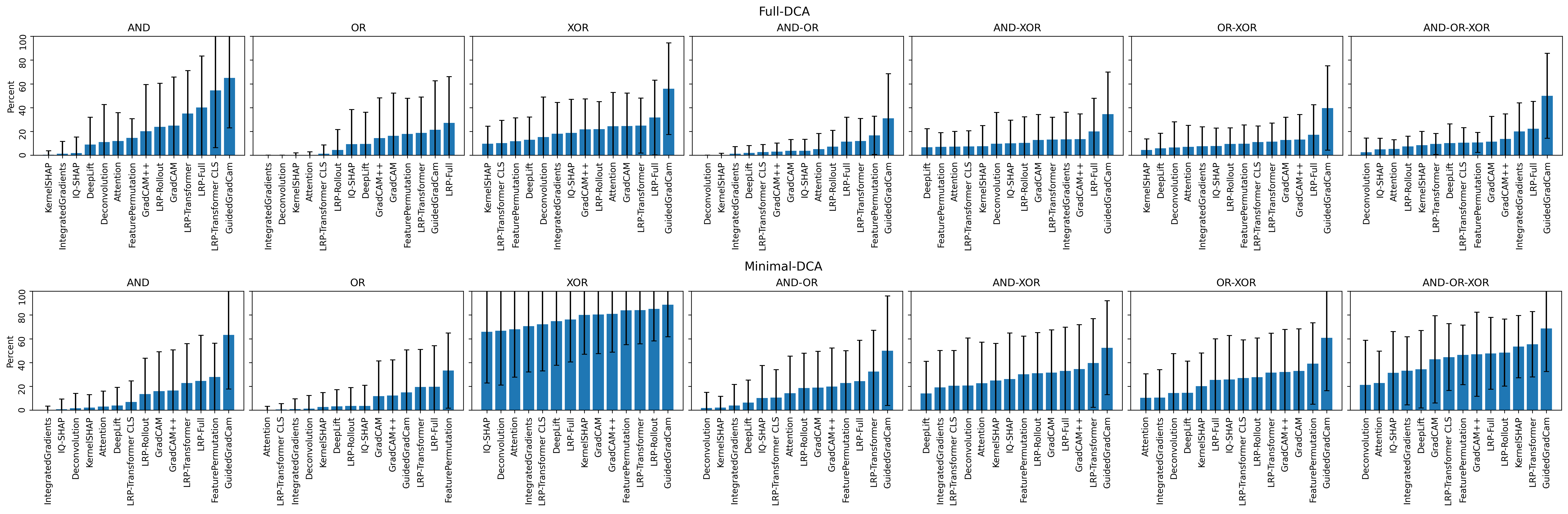}
	\caption{verage Full-DCA and minimal-DCA results for each saliency method, ranked in and differentiated between the different scenarios. Considering all trained base models.}
	\label{fig:dcaFull}
\end{figure}

\begin{figure}[ht!]
	\centering
	\includegraphics[width=0.99\columnwidth]{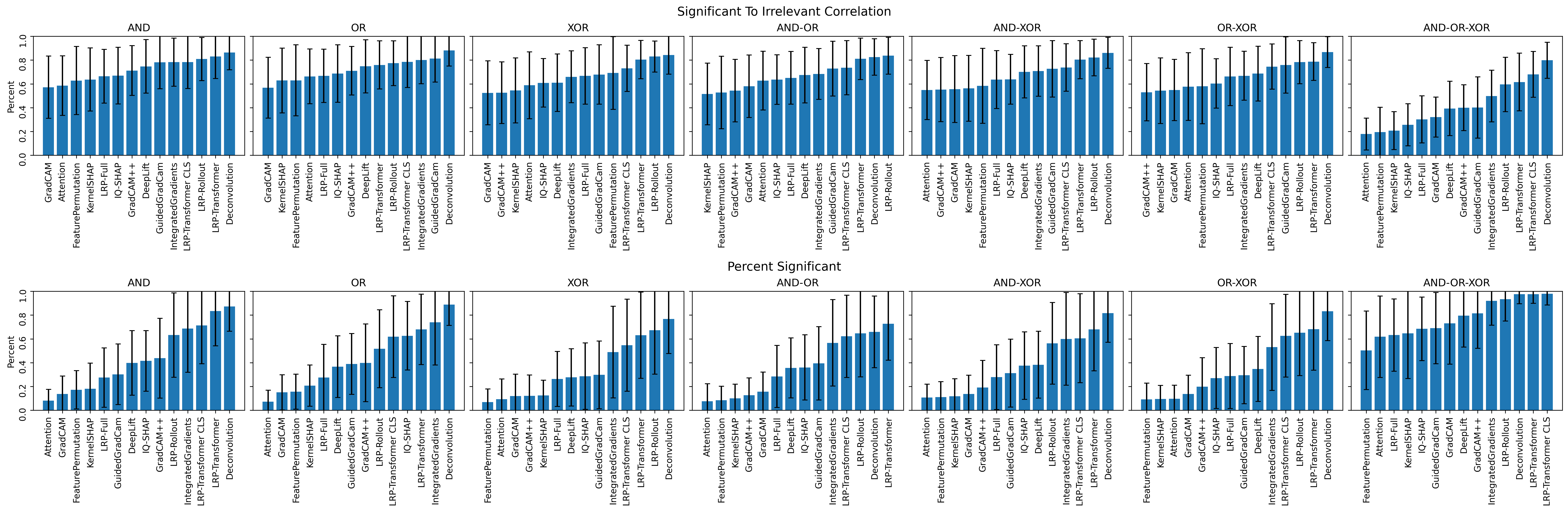}
	\caption{Average significant (p $<$ 0.5) Pearson correlation from the baseline scores to all other scores from the sample, averaged over all models (top) and
average (over all models) percentage of correlations that are significant from all possible ones (bottom). Results ranked in and differentiated between the
different scenarios.}
	\label{fig:correlationFull}
\end{figure}

\begin{figure}[ht!]
	\centering
	\includegraphics[width=0.99\columnwidth]{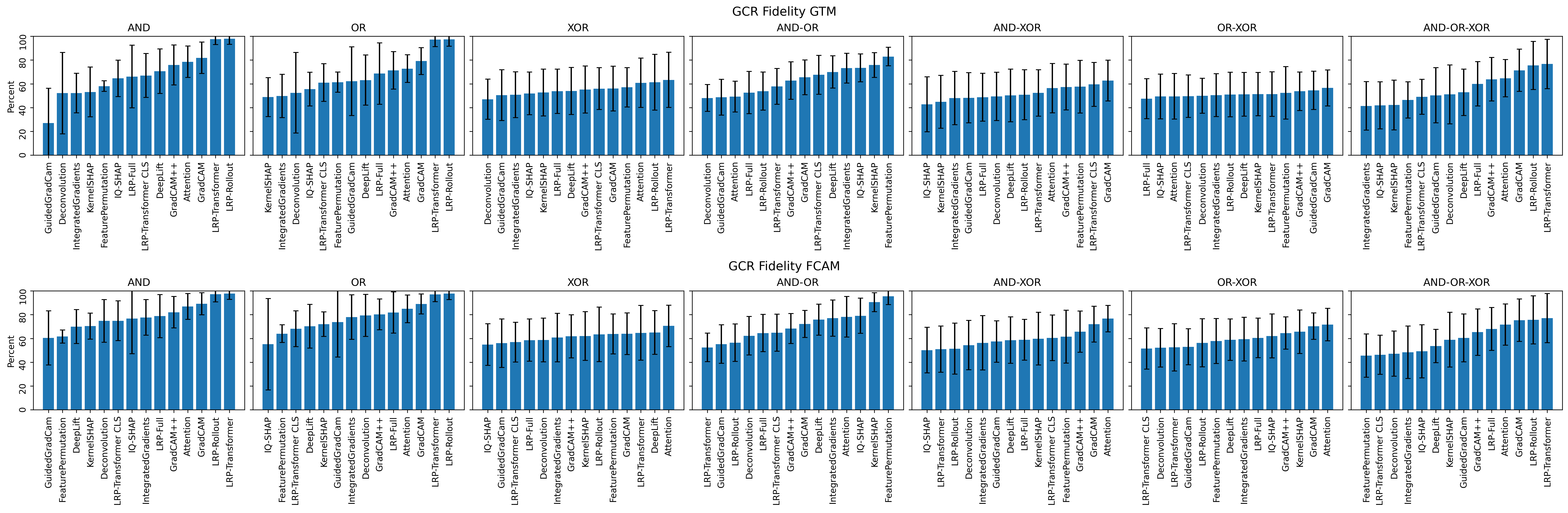}
	\caption{Average GCR fidelity results for each saliency method, ranked in and differentiated between the different scenarios. Considering all trained base models.}
	\label{fig:gcrFull}
\end{figure}

\begin{figure}[ht!]
	\centering
	\includegraphics[width=0.99\columnwidth]{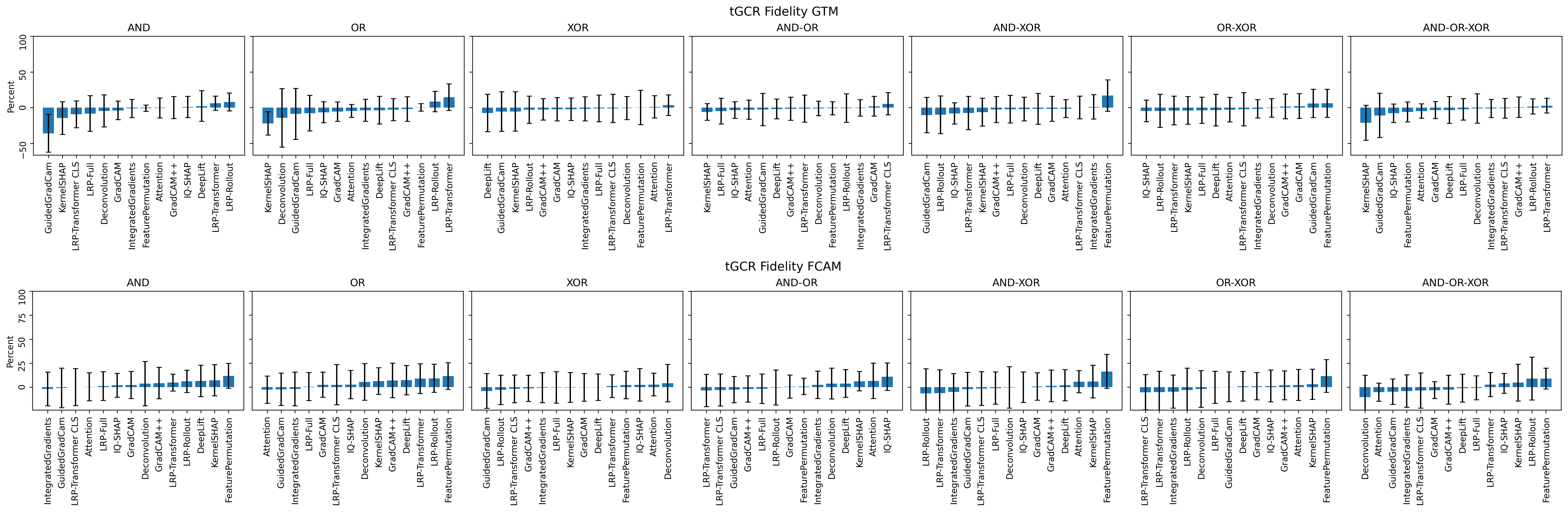}
	\caption{Average tGCR fidelity results for each saliency method, ranked in and differentiated between the different scenarios. Considering all trained base models.}
	\label{fig:tgcrFull}
\end{figure}

\end{document}